%% file: main.tex
\documentclass[10pt,twocolumn,letterpaper]{article}

\usepackage{cvpr}      
\usepackage[accsupp]{axessibility} 
\usepackage[ruled,linesnumbered]{algorithm2e}
\usepackage{amsmath,amsfonts}
\usepackage{algorithmic}
\usepackage{array}
\usepackage{textcomp}
\usepackage{stfloats}
\usepackage{url}
\usepackage{verbatim}
\usepackage{graphicx}
\usepackage{amsmath}
\usepackage{amssymb}
\usepackage{booktabs}
\usepackage{stfloats}

\usepackage[utf8]{inputenc} 
\usepackage[T1]{fontenc}    
\usepackage{url}            
\usepackage{booktabs}       
\usepackage{amsfonts}       
\usepackage{nicefrac}       
\usepackage{microtype}      
\usepackage{xcolor}         
\usepackage{multirow}
\usepackage{xspace}
\usepackage{float}


\definecolor{cvprblue}{rgb}{0.21,0.49,0.74}
\usepackage[pagebackref,breaklinks,colorlinks,citecolor=cvprblue]{hyperref}

\def\paperID{5340} 
\def\confName{CVPR}
\def\confYear{2024}

\newcommand{\hzx}{\textcolor{black}}
\def\eg{\emph{e.g.,}} 
\def\ie{\emph{i.e.,}} 
\def\cf{\emph{c.f.}} 
 
\def\wrt{{w.r.t.}} 
\def\sexyname{G-NeRF\xspace}
\definecolor{gray}{RGB}{100,100,100}
\def\BibTeX{{\rm B\kern-.05em{\sc i\kern-.025em b}\kern-.08em
    T\kern-.1667em\lower.7ex\hbox{E}\kern-.125emX}}
\usepackage{balance}
\usepackage{pifont}
\newcommand{\cmark}{\ding{51}}%
\newcommand{\xmark}{\ding{55}}%
\title{G-NeRF: Geometry-enhanced Novel View Synthesis from Single-View Images}
\author{Zixiong Huang$^{1, 3}$\footnotemark[1]~~~~Qi Chen$^{2}$\footnotemark[1]~~~~Libo Sun$^2$~~Yifan Yang$^1$~~~\\
Naizhou Wang$^3$~~~Mingkui Tan$^{1, 4, 5}$\footnotemark[2]~~~Qi Wu$^2$\\
$^1$South China University of Technology~~$^2$The University of Adelaide~~\\
$^3$Guangzhou Shiyuan Electronics Co., Ltd~~$^4$Pazhou Lab\\
$^5$Key Laboratory of Big Data and Intelligent Robot, Ministry of Education\\
{\tt\small sesmilhzx@mail.scut.edu.cn, qi.chen04@adelaide.edu.au; mingkuitan@scut.edu.cn}  \\
}

\begin{document}
\maketitle
\renewcommand{\thefootnote}{\fnsymbol{footnote}}
\footnotetext[2]{Corresponding author, *Authors contributed equally.}
\footnotetext{Project page: \href{https://llrtt.github.io/G-NeRF-Demo/}{https://llrtt.github.io/G-NeRF-Demo/}}
\footnotetext{This work was done when Zixiong Huang was a research intern at Guangzhou Shiyuan Electronics Co., Ltd.}
\renewcommand{\thefootnote}{\arabic{footnote}}
\input{sec/0_abstract}    
\input{sec/1_intro}
\input{sec/2_formatting}

{
    \small
    \bibliographystyle{ieeenat_fullname}
    \bibliography{main}
}

\clearpage
\maketitlesupplementary
\input{sec/X_suppl}

\end{document}

%% file: sec/0_abstract.tex
\begin{abstract}
Novel view synthesis aims to generate new view images of a given view image collection.
Recent attempts address this problem relying on 3D geometry priors~(\eg~shapes, sizes, and positions) learned from multi-view images. However, such methods encounter the following limitations: 1)~they require a set of multi-view images as training data for a specific scene~(\eg~face, car or chair), which is often unavailable in many real-world scenarios;
2)~they fail to extract the geometry priors from single-view images due to the lack of multi-view supervision.
In this paper, we propose a Geometry-enhanced NeRF (G-NeRF), which seeks to enhance the geometry priors by a geometry-guided multi-view synthesis approach, followed by a depth-aware training.
In the synthesis process, inspired that existing 3D~GAN models can unconditionally synthesize high-fidelity multi-view images, we seek to adopt off-the-shelf 3D~GAN models, such as EG3D, as a free source to provide geometry priors through synthesizing multi-view data.
Simultaneously, to further improve the geometry quality of the synthetic data, we introduce a truncation method to effectively sample latent codes within 3D~GAN models.
To tackle the absence of multi-view supervision for single-view images, we design the depth-aware training approach, incorporating a depth-aware discriminator to guide geometry priors through depth maps.
Experiments demonstrate the effectiveness of our method in terms of both qualitative and quantitative results.

\end{abstract}

%% file: sec/1_intro.tex
\section{Introduction}

\begin{figure}[t]
    \centering
    \includegraphics[width=1.0\linewidth]{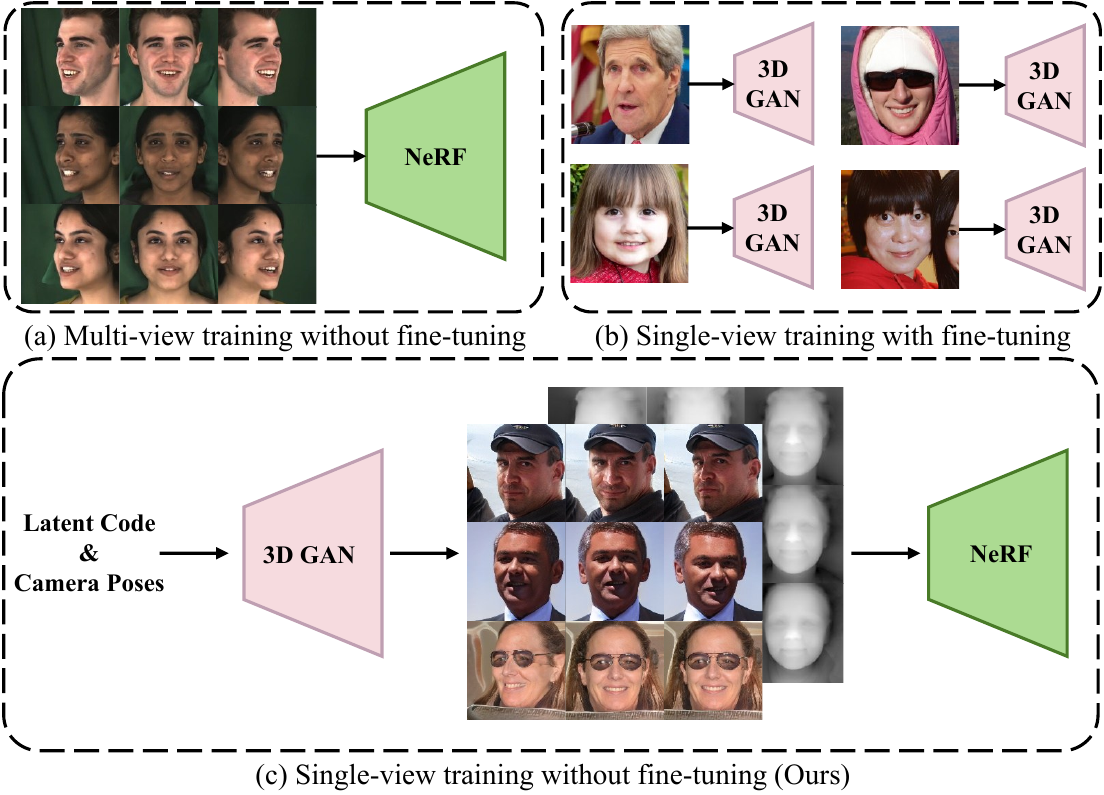}
    \caption{\textbf{Comparison of different methods.}
            To achieve single-shot novel view synthesis, previous methods either (a) require real-world multi-view images to establish geometry priors or (b) need additional optimization for a specific image.
            (c) In contrast, our method captures the geometry priors from an existing 3D GAN trained on single-view images only.}
    \label{introduction}
    \vspace{-20pt}
\end{figure}

\label{sec:intro}
Neural Radiance Fields (NeRFs) have emerged as a state-of-the-art technique for synthesizing novel views of complex scenes from 2D images. By using deep neural networks, NeRF models are capable of modeling both the geometry and appearance of a scene, enabling the generation of high-quality and photorealistic 3D renderings from any desired viewpoint.
Its remarkable performance has made it a mainstream method for novel view synthesis, and it has found diverse applications in fields such as virtual reality and digital human generation~\cite{guo2021ad, deng2022fov}.
Although NeRF has demonstrated exceptional performance in synthesizing novel views for many scenes, it does exhibit two prominent limitations.

First, it requires multi-view images for training on a specific scene, while in most practical conditions only a single-view image is available.
This limitation is widespread in real-world scenarios, such as when taking selfies or capturing a portrait of a pet.
Some recent works~\cite{xu2022sinnerf, deng2022depth, yu2021pixelnerf} have attempted to address this limitation by introducing additional supervision, such as depth, or by collecting extensive multi-view images of the same class to learn sufficient geometry priors.
For example, 
SinNeRF~\cite{xu2022sinnerf} adopts
ground truth depth or depth obtained from multi-view images to train a NeRF model for each individual image. 
Nonetheless, it requires accurate depth information, which is often hard to obtain from a single image.
PixelNeRF~\cite{yu2021pixelnerf} can perform few-shot or even single-shot novel view synthesis by collecting a set of multi-view images of the same class for training. However, it is inapplicable to some real-world scenarios where only single-view images are accessible.
\hzx{Furthermore, Pix2NeRF directly optimizes with image-level reconstruction and GAN losses.
This, however, may incur a challenging one-to-many problem, as there could be multiple 3D shapes corresponding to one input image~(\cf~Sec 4.4).}


Second, the conventional NeRF model only focuses on overfitting to a particular scene while ignoring the sharable intrinsic geometry prior among the relevant scenes, such as scenes from the same class~(\eg~faces, cars or chairs).
To address this issue, a recent line of research~\cite{pavllo2023shape, rematas2021sharf, chan2022efficient, gu2022stylenerf} has explored the combination of Generative Adversarial Networks (GANs) \cite{goodfellow2014generative} and NeRF to extract 3D prior knowledge from single-view datasets like FFHQ~\cite{karras2019style} and AFHQv2~\cite{choi2020stargan}.
While successful in diverse high-quality 3D scene generation, these models synthesize images from randomly sampled latent codes.
This means to generate novel views for a specific image, we have to first map the image back to the latent space to obtain a corresponding latent code, which leads to extra test-time optimizations (\eg~GAN inversion~\cite{roich2021pivotal}) with additional time consumption. 
Additionally, these approaches adopt conventional reconstruction loss to fine-tune 3D GANs, neglecting the intrinsic geometry information embedded in these models. Thus, they tend to yield suboptimal outcomes characterized by geometry collapse~\cite{yuan2023make}, primarily due to the absence of multi-view supervision. 

To address the above issues, we seek to develop a framework that can accomplish two goals simultaneously: exploiting geometry priors from an existing 3D GAN trained on real-world single-view images only and enabling single-shot novel view synthesis without the need for test-time fine-tuning~(see Fig.\ref{introduction}).
To achieve this, we propose a novel single-shot novel view synthesis approach named  Geometry-enhanced NeRF~(\sexyname), which seeks to enhance the geometry priors by two approaches: Geometry-guided Multi-View Synthesis~(GMVS) and Depth-aware Training (DaT). 
In GMVS phase, we use a pre-trained 3D GAN model to generate a set of multi-view data, serving as a free source for establishing geometry priors. 
To further enhance the geometry priors in the synthetic data, we trade off the diversity against the geometry quality in 3D GANs, exploring different truncation ratios~\cite{brock2019large} to achieve a balance for high-quality geometry data synthesis.
In the DaT training phase, we introduce a depth-aware discriminator to address the lack of multi-view supervision for single-view images. This discriminator distinguishes between the depth maps generated by the pre-trained 3D GAN model and those produced by our model, thereby providing additional depth supervision to enhance the geometry fidelity of our generated results.
We summarize our contributions in three folds:
\begin{itemize}
    \item To obtain sufficient multi-view images for training a single-shot NeRF model, we propose a Geometry-guided Multi-View Synthesis scheme to synthesize a set of multi-view data to build adequate geometry priors.
    \item To generate high-quality synthetic data, we explore the trade-off between diversity and geometry quality in 3D GANs, and then introduce a truncation method for multi-view data synthesis with enhanced geometry priors.
    \item For better learning geometry priors from single-view images, we design a Depth-aware Training method. It adots a depth-aware discriminator to enhance depth supervision, guiding the model to generate more realistic results.
\end{itemize}

%% file: sec/2_formatting.tex
\vspace{-7pt}
\section{Related Work}

\noindent\textbf{Neural Radiance Fields for few views.}
Representing 3D scenes as an implicit MLP-based function and using volume rendering technology, NeRF~\cite{mildenhall2020nerf} and its variants~\cite{barron2021mip,zhang2020nerf++,muller2022instant,wang2021nerf,pumarola2021d, tang2022compressible, yang2023cross} have shown promising results in novel view synthesis tasks.
However, these methods face a common challenge: they require a large number of views to obtain sufficient density information.

To address the data-hungry nature of NeRF, recent works~\cite{xu2022sinnerf,yu2021pixelnerf,deng2022depth, chen2021mvsnerf, yang2022s3nerf, shen2023gina} aim to learn shared priors or incorporate additional supervision, such as depth maps.
PixelNeRF~\cite{yu2021pixelnerf} conditions NeRF on images by computing a fully convolutional image feature, which serves as a 3D representation for volumetric rendering. This approach allows for predicting NeRF from images in a feed-forward manner while leveraging shared priors of scenarios.
DS-NeRF~\cite{deng2022depth} introduces additional supervision by leveraging depth information recovered from 3D point clouds estimated using structure-from-motion methods.
SinNeRF~\cite{xu2022sinnerf} further divides the NeRF training process into geometry learning and semantic learning.
Combining camera and LiDAR data, GINA-3D~\cite{shen2023gina} achieves generating neural assets from a single-view input.
Despite their advancements, these methods still rely on a collection of multi-view images to provide an adequate 3D prior or require additional depth supervision~(\eg, LiDAR data, depth map).

\vspace{5pt}
\noindent\textbf{Generative 3D-aware image synthesis.}
Generative Adversarial Nets (GANs)~\cite{goodfellow2014gans, karras2019style, mirza2014conditional, song2021agegan++, han2019asymmetric, li2022learning, guo2019auto} have achieved impressive success in 2D image synthesis tasks~(\eg~image generation, image-to-image translation).
Recently, many attempts~\cite{chan2021pi, schwarz2020graf, niemeyer2021giraffe, cai2022pix2nerf, nguyen2019hologan, gu2022stylenerf, chan2022efficient} have been made to extend GANs to 3D-aware tasks.
HoloGAN~\cite{nguyen2019hologan}successfully disentangles 3D pose and identity by using an explicit volume representation, but this type of 3D representation also limits the resolution of the generated images.
Some methods~\cite{chan2021pi, schwarz2020graf, niemeyer2021giraffe, chan2022efficient,gu2022stylenerf, deng2022gram} combine GANs and NeRF to synthesize high-fidelity novel views.
However, these methods generate random scenes using randomly sampled latent codes. 
To achieve single-view reconstruction, additional test-time optimization~(\eg~GAN inversion) is necessary~\cite{yin2022nerfinvertor, lan2023self, yuan2023make}.
While these methods excel at producing high-fidelity novel perspective images, they are constrained to the original 3D GAN model. Essentially, they rely on the original model for every inference. In contrast, our approach is versatile and capable of being applied to any model of the same type.
\begin{figure*}[ht]
    \centering
    \includegraphics[width=1.9\columnwidth]{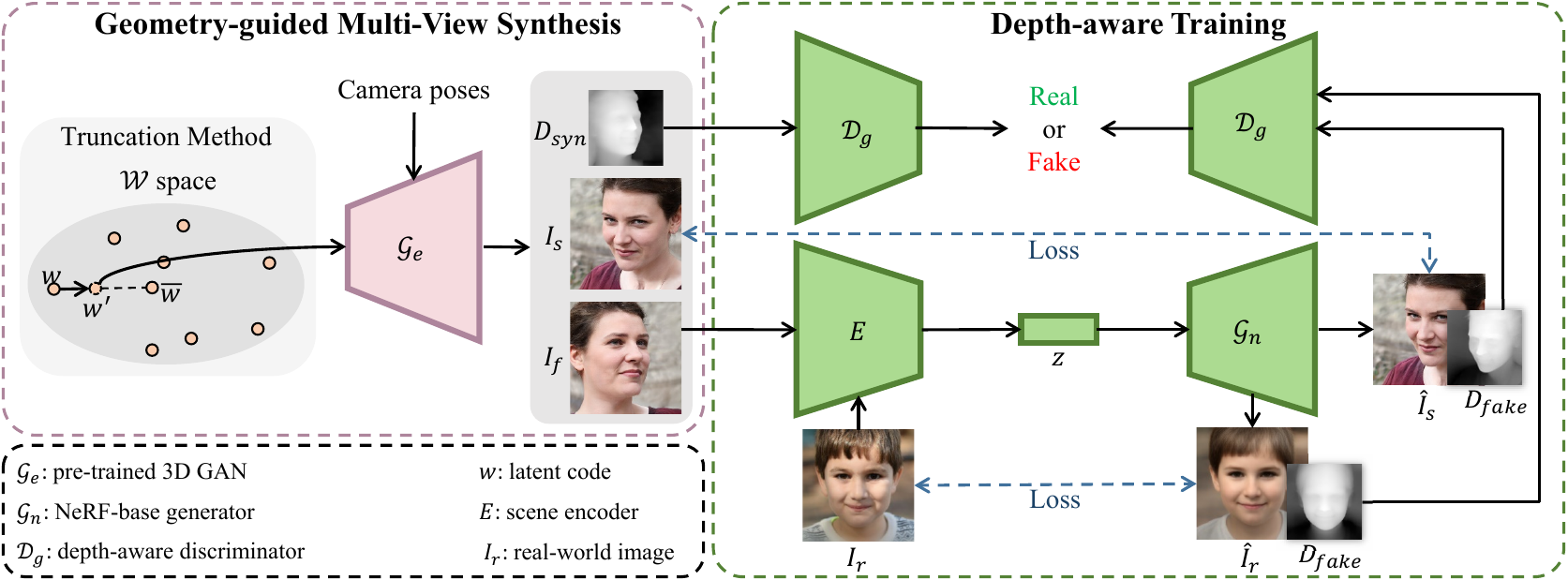}
    \caption{\textbf{Overall scheme of \sexyname.}
    Given a latent code $\mathbf{w}$ randomly sample in $\mathcal{W}$ space, we first apply a truncation method to obtain $\mathbf{w}^{\prime}$, bringing it closer to the center of mass of $\mathcal{W}$ space represented $\bar{\mathbf{w}}$. After that, in conjunction with a set of camera poses $\{\mathbf{P}_{f}, \mathbf{P}_{s}, \mathbf{P}_{d}\}$, we generate a triplet of synthetic data $\{{I}_{f}, {I}_{s}, \mathbf{D}_{syn}\}$.
    To capture geometry priors from synthetic multi-view images, we synthesize a novel view $\hat{I}_{s}$ using ${I}_{f}$ as the reference image and enforce it to be consistent with ${I}_{s}$.
    Additionally, we employ a self-reconstruction task with the real-world image ${I}{r}$.
    Moreover, we design a depth-aware discriminator $\mathcal{D}_{g}$ to further enhance the depth quality of the generated scenes.}
    \label{fig: overallarchitecture}
    \vspace{-10pt}
\end{figure*}

\section{Geomery-enhanced NeRF}
We aim to address the task of single-shot novel view synthesis in a unified framework. In other words, given an unseen single-view image~(\eg~human face or cat face), our goal is to synthesize novel views of the same scene.
This task is inherently difficult due to the limited geometry information available in a single-view image. 
One potential solution involves leveraging multi-view datasets to establish adequate geometry priors.
However, obtaining such datasets may be impractical in many real-world scenarios.
To overcome this limitation, we introduce Geometry-enhanced NeRF (\sexyname), a novel approach designed to achieve high-fidelity single-shot novel synthesis from single-view images.

As shown in Fig.~\ref{fig: overallarchitecture}, \sexyname consists of two stages:~1) \textit{Geometry-guided multi-view synthesis}~(\cf~Sec.~\ref{multi-view dataset generation}). 
To learn geometry priors of similar scenes, we leverage a pre-trained 3D GAN model $\mathcal{G}_{e}$ to synthesize a collection of multi-view images and corresponding depth maps.
Simultaneously, we delve into the trade-off between diversity and geometry quality in 3D GANs, 
proposing an approach
to achieve a balance for Geometry-guided data synthesis.
2) \textit{Depth-aware training.}~(\cf~Sec.~\ref{training}).
We seek to train our model using the combination of synthetic data and real-world single-view images with a reconstruction loss, denoted as $\mathcal{L}_{recon}$.
However, solely applying $\mathcal{L}_{recon}$ is insufficient to learn satisfying geometry priors due to the absence of multi-view supervision for single-view images.
To provide additional supervision, we introduce a depth-aware discriminator, which helps distinguish between the synthetic depth maps generated by our model and the depth maps obtained from the pre-trained 3D GAN model. We incorporate it into our training through a depth-aware adversarial loss, denoted as $\mathcal{L}{gan}$.

The overall optimization of our proposed \sexyname minimizes the following objective function:
\begin{equation}
    \begin{aligned}
     \mathcal{L}_{total} = \mathcal{L}_{recon} + \lambda_{g}\mathcal{L}_{gan},
    \label{eq: total}
    \end{aligned}
\end{equation}
where $\lambda_{g}$ is a hyper-parameter used to balance the reconstruction loss $\mathcal{L}_{recon}$~(see Eqn.~(\ref{eq: recon loss})) and the adversarial loss $\mathcal{L}_{gan}$~(see Eqn.~(\ref{eq: disc loss})).

\subsection{Geometry-guided Multi-View Synthesis}
In the absence of multi-view supervision, it is challenging to learn geometry information from single-view datasets, such as FFHQ~\cite{karras2019style} and AFHQv2-Cats~\cite{choi2020stargan}.
Inspired that existing 3D GAN models can synthesize high-fidelity multi-view images while being trained solely on a set of single-view images.
We aim to leverage the rich geometry priors embedded in these models.
To this end, we propose a Geometry-guided Multi-View Synthesis scheme.
In GMVS, we first utilize an off-the-shelf 3D GAN~(\eg~EG3D~\cite{chan2022efficient}) to synthesize a set of multi-view data.
However, the use of naive synthetic data can lead to suboptimal results, such as unrealistic 3D shapes~(see Fig.~\ref{fig: Ablation Studies of GMVS}).
To address this, we delve into the balance between diversity and geometry quality within 3D GANs.
Building upon this exploration, we apply a latent truncation method to strike a more suitable balance between the diversity and geometry quality of the synthetic data, thereby contributing to the generation of more realistic results.

\vspace{-8pt}
\paragraph{Trade-off between diversity and geometry quality.}
\label{Diversity and Fidelity Exploration Scheme}
As confirmed by previous studies~\cite{brock2019large,alemohammad2023self}, a trade-off exists between the fidelity and diversity of samples generated by GAN models.
Take StyleGAN-based~\cite{karras2019style} methods as an example, we randomly sample a latent code $\mathbf{z}_{d}  \sim p_z \subset \mathbb{R}^{512}$, where $p_z$ is a normal distribution.
Then, a mapping network $\mathcal{M}$ is adopted to map $\mathbf{z}_{d}$ to an intermediate latent space $\mathcal{W}$ to acquire $\mathbf{w}$.
After that, as illustrated in Fig.~\ref{fig: overallarchitecture}, a truncation method is leveraged to draw $\mathbf{w}$ closer to the center of mass of $\mathcal{W}$ space by:
\begin{equation}
    \begin{aligned}
     \mathbf{w}^{\prime} = \bar{\mathbf{w}} + \psi(\mathbf{w}-\bar{\mathbf{w}}),
    \label{eq: truncation trick}
    \end{aligned}
\end{equation}
where $\bar{\mathbf{w}}=\mathbb{E}_{\mathbf{z}  \sim p_z}[\mathcal{M}(\mathbf{z})]$ is the center of mass of $\mathcal{W}$ and $\psi \leq 1$ is a truncation ratio.
The manipulation of $\psi$ allows us to finely tune the trade-off between diversity and fidelity.
Specifically, an increase in $\psi$ augments diversity but may simultaneously diminish fidelity or the visual appeal of the generated results.
This adjustment is driven by the fact that regions with lower density may be inadequately represented, posing challenges for the generator to effectively learn.

\begin{figure}[t]
    \centering
    \subfloat[$\psi=0.0$]{\includegraphics[width=0.48\columnwidth]{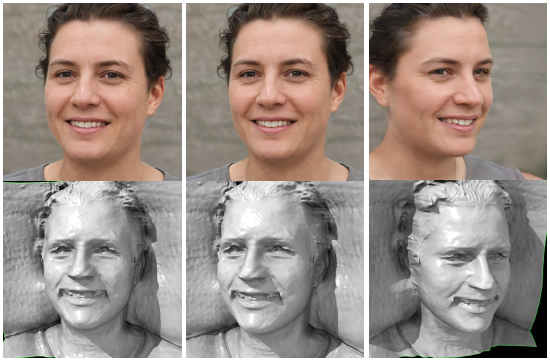}
     \label{truncat_0.0}}
    \subfloat[$\psi=0.3$]{\includegraphics[width=0.48\columnwidth]{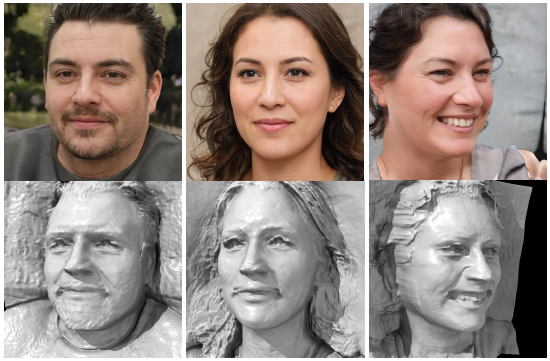}
     \label{truncat_0.3}}
     
    \subfloat[$\psi=0.7$]{\includegraphics[width=0.48\columnwidth]{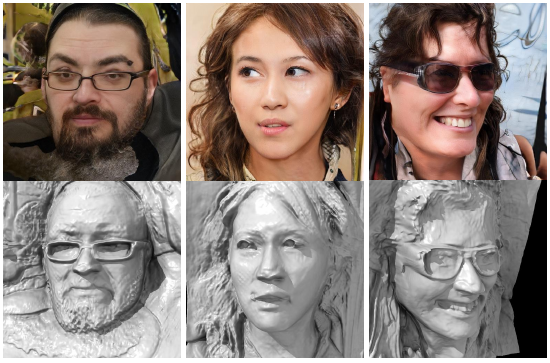}
     \label{truncat_0.7}}
    \subfloat[$\psi=1.0$]{\includegraphics[width=0.48\columnwidth]{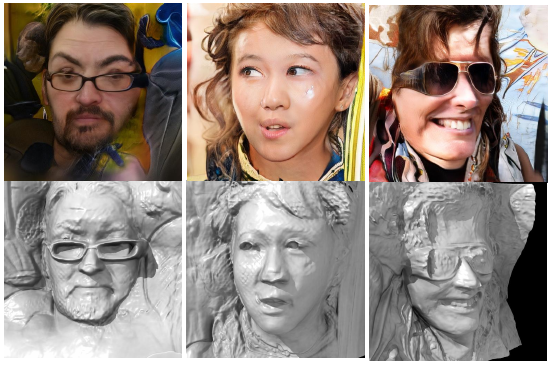}
     \label{truncat_1.0}}
    \vspace{-2pt}
    \caption{Illustration of the trade-off between identity diversity and geometry quality of the generated samples.
    Samples are generated by EG3D~\cite{chan2022efficient} with the same set of latent codes and different truncation ratios $\psi$.
    As $\psi$ rises, the identity diversity~(\eg~hair color, skin color, and glasses) of the generated samples also increases.
    In contrast, the geometry quality of these scenes gradually reduces.
    }
    \label{fig: truncation effect}
    \vspace{-8pt}
\end{figure}

In this study, we delve deeper into this phenomenon within the realm of 3D GANs. 
Specifically, we leverage EG3D~\cite{chan2022efficient} to generate four sets of samples with varying truncation ratios and examine their differences. 
As depicted in Fig.\ref{fig: truncation effect}, our results demonstrate an augmented diversity with increasing truncation ratios, albeit accompanied by a gradual reduction in geometry quality. 

\vspace{-10pt}
\paragraph{Geometry-guided multi-view synthesis.}
\label{multi-view dataset generation}
Building upon the above observation, we conclude that synthetic data generated with various truncation ratios plays a crucial role in the geometry quality of final results. To obtain a set of multi-view data with high-quality geometry priors, we devise a Geometry-guided Multi-View Synthesis scheme. 
Specifically, we employ a pre-trained 3D GAN model $\mathcal{G}_{e}$, such as EG3D~\cite{chan2022efficient}, to generate multi-view data of diverse scenes. 
As shown in Fig.~\ref{fig: overallarchitecture}, we randomly sample a latent code $\mathbf{w}$ and a set of camera pose $(\mathbf{P}_{f},\mathbf{P}_{s},\mathbf{P}_{d}) \sim p_{\xi}$, where $p_{\xi}$ is a distribution associated with camera poses from real-world single-view images. 
We then apply the truncation method with an empirically selected ratio $\psi=0.5$ to obtain a truncated latent code $\mathbf{w}^{\prime}$ by Eqn.~(\ref{eq: truncation trick}).
As discussed earlier, this truncation method draws $\mathbf{w}$ closer to the center of mass of $\mathcal{W}$ space, ensuring the geometry quality of the generated results.
Finally, we synthesize a triplet of Geometry-guided multi-view data $\mathbf{I}$ with a generator $\mathcal{G}_{e}$ by:
\begin{equation}
\label{eq: data synthesis}
    \begin{split}
        \mathbf{I}&=\{{I}_{f}, {I}_{s}, \textbf{D}_{syn}\}=\mathcal{G}_{e}({\mathbf{w}}^{\prime}, \mathbf{P}_{f}, \mathbf{P}_{s}, \mathbf{P}_{d}),
    \end{split}
\end{equation}
where ${{I}}_{f}$ and ${{I}}_{s}$ denote the first and second 
synthetic images regarding a common scene but rendered from different viewpoints, and ${\mathbf{D}}_{syn}$ denotes the depth map of the scene.

\subsection{Depth-aware Training}
\label{training}
In this section, we seek to train our model with a combination of synthetic data and real-world single-view images.
However, directly applying reconstruction loss to single-view images is not beneficial for learning satisfying geometry priors due to the absence of multi-view supervision.
To address this, we introduce a depth-aware discriminator $\mathcal{D}_{g}$ to provide additional depth supervision.

\vspace{-5pt}
\paragraph{Incorporating synthetic and real-world data.}
We incorporate synthetic data with real-world single-view images to train our model.
Specifically, we simultaneously generate a novel view and depth map with a selection factor $\gamma \sim \mathcal{U}(0,1)$, which is formulated as:
\begin{equation}
    \begin{split}
            \begin{aligned}
                \hat{I}_{s},\mathbf{D}_{fake} = &\mathcal{G}_{n}(\mathbf{P}_{s}, {E}({I}_{f})), &&\text{if } 0 \leq \gamma \leq 0.5; \\
                \hat{I}_{r},\mathbf{D}_{fake} = &\mathcal{G}_{n}(\mathbf{P}_{r}, {E}({I}_{r})), &&\text{if } 0.5 < \gamma \leq 1,
            \end{aligned}
    \label{eq: domain mixture} 
    \end{split}
\end{equation}
where $I_r$ is a real-world image associated with the pose $\mathbf{P}_r$, $\mathcal{G}_n$ is a NeRF-based generator and ${E}$ is an scene encoder.
This alternative training scheme allows the model to capture the geometry priors within the multi-view images, while still learning diverse appearance information from the real-world images.
With the generated views and depth maps, we depict a reconstruction loss and an adversarial loss as follows. 

\vspace{-8pt}
\paragraph{Reconstruction with paired images.} For synthetic multi-view image pairs $\{I_f, I_s\}$ , we adopt $I_f$ as the reference image and render a novel view image $\hat{I}_s$ from the same viewpoint as $I_s$. 
In this way, we train our \sexyname by minimizing photometric error \wrt~$I_s$ and $\hat{I}_s$.
For real-world single-view images, our objective is to leverage them to augment the diversity of the synthesized scenes.
To this end, we select a single-view image $I_r$ as a reference and train \sexyname by generating a $\hat{I}_{r}$ that shares the same viewpoint with $I_r$. We update \sexyname by enforcing similarity between $\hat{I}_{r}$ and ${I}_{r}$. 
This schedule is implemented using a reconstruction loss, which is formulated with the image-pair data $({I}_{fake}, {I}_{ref}) {\in} \{(\hat{I}_s, I_s ), (\hat{I}_r, I_r)\}$, 
\begin{equation}
    \label{eq: recon loss}
    \begin{aligned}
     \mathcal{L}_{recon} &= \mathbb{E}\left[||{I}_{fake}-{I}_{ref}||_{1} + \mathcal{L}_{ssim}({I}_{fake}, {I}_{ref})\right. \\
     &\left. + \mathcal{L}_{vgg}({I}_{fake}, {I}_{ref})\right]. 
    \end{aligned}
\end{equation}
$\mathcal{L}_{ssim}$ is SSIM loss~\cite{wang2004image} and $\mathcal{L}_{vgg}$ is perceptual loss~\cite{johnson2016perceptual}.

\begin{figure*}[t]
    \centering
    \includegraphics[width=0.9\linewidth]{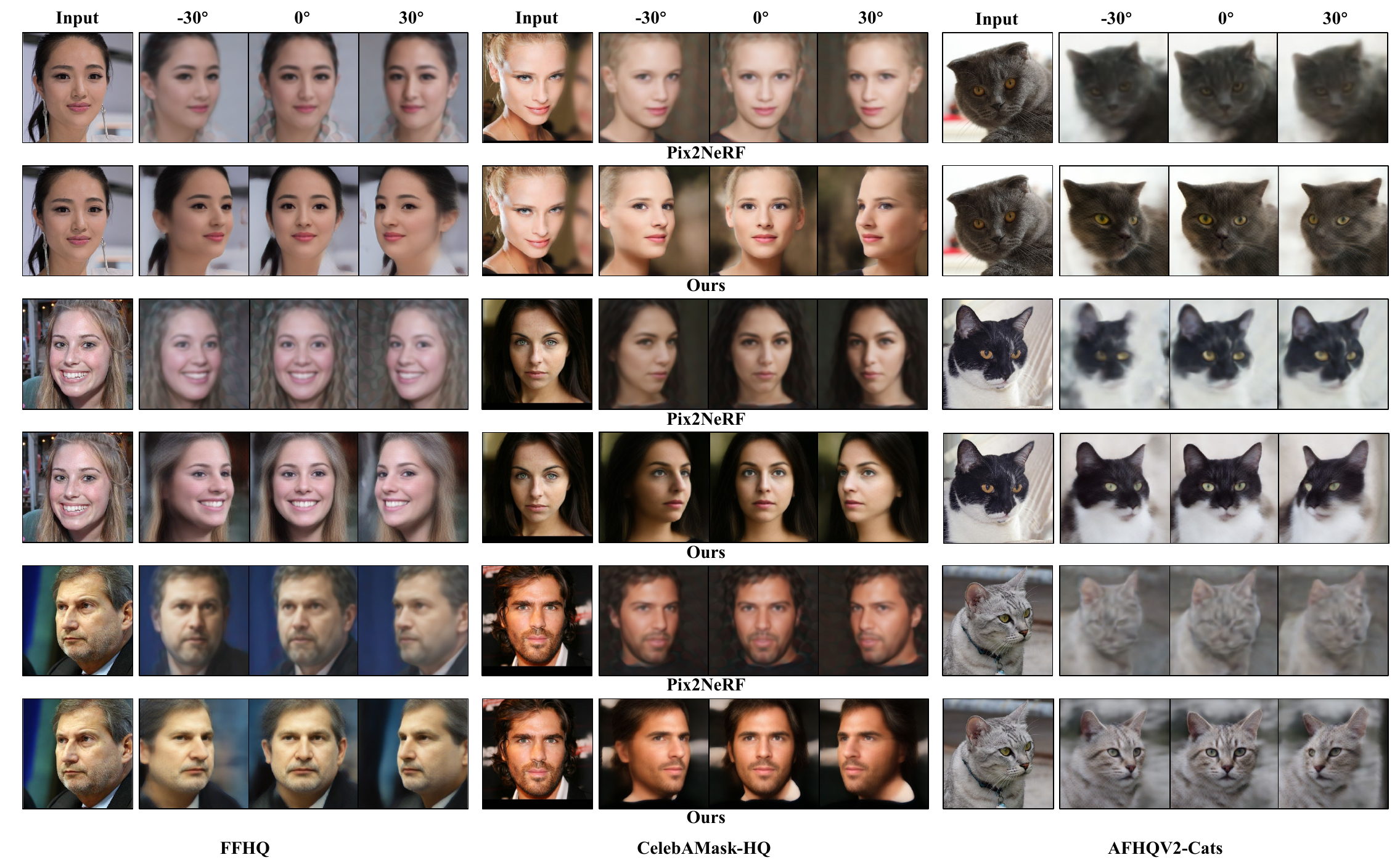}
    \vspace{-10pt}
    \caption{\textbf{Qualitative comparison.}
    Compared to Pix2NeRF~\cite{cai2022pix2nerf}, our \sexyname demonstrates the capability to generate novel views that closely resemble reference images with higher clarity~(Comparison at $512^2$).}
    \label{fig: Qualitative Comparison}
    \vspace{-15pt}
\end{figure*}


\paragraph{Depth-aware discriminator.}
\label{Depth-aware discriminator}
In the absence of multi-view supervision for real-world single-view images, we observe some degradation in the geometry quality of our generated results~(see Fig.~\ref{fig: Ablation Studies of GMVS}).
To address this, we introduce a depth-aware discriminator denoted as $\mathcal{D}_{g}$.
Concretely,  $\mathcal{D}_{g}$ is trained to distinguish between the generated depth map ${\mathbf{D}}_{fake}$ and ground truth $\mathbf{D}_{syn}$ from the synthetic data, thereby introduce additional geometry priors into our model.
In contrast to employing a simple reconstruction loss for basic depth supervision, $\mathcal{D}_{g}$ offers several advantages: 1) enabling depth supervision for real-world single-view images whose depth maps may not be available; 2) ensuring that depth maps generated from various viewpoints are realistic and coherent with the scene, contributing to the overall quality of the synthesized novel views.
Following EG3D~\cite{chan2022efficient}, we condition $\mathcal{D}_{g}$ on a camera pose and use an adversarial loss with an R1 regularization~\cite{mescheder2018training} to train $\mathcal{D}_{g}$:
\begin{equation}
    \begin{aligned}
     \mathcal{L}_{gan} =~& \mathbb{E}[f(\mathcal{D}_{g}(\textbf{D}_{syn}|\mathbf{P}_{d})] \\
     +& \mathbb{E}\left[f(-\mathcal{D}_{g}({\textbf{D}}_{fake}|\mathbf{P}_{f}))+\lambda|\nabla {\mathcal{D}_{g}({\textbf{D}}_{fake}}|\mathbf{P}_{f})|^2\right], 
    \label{eq: disc loss} 
    \end{aligned}
\end{equation}
where $\mathbf{P}_{f} {\in} (\mathbf{P}_{r}, \mathbf{P}_{s})$ is the camera pose used to generate novel views and $f(\cdot)$ is a softplus activation.
Note that, for each real-world image, we train $\mathcal{D}_g$ not only with its corresponding camera pose $\mathbf{P}_{r}$ but also with other camera poses randomly sampled from ${p}_{\xi}$ to provide more comprehensive depth supervision.

\section{Experiments}

\subsection{Experimental Setup}
\paragraph{Datasets.}
We train our model with FFHQ~\cite{karras2019style} and AFHQv2-Cats~\cite{choi2020stargan}, repectively.
During the evaluation, we leverage an additional in-the-wild dataset named CelebAMask-HQ~\cite{lee2020maskgan}.
FFHQ~\cite{karras2019style} is a real-world dataset with around 70k high-quality human faces.
CelebAMask-HQ~\cite{lee2020maskgan} is a large-scale face dataset with 30k high-resolution human faces.
After preprocessing, we randomly hold out 8k images as the test set. 
AFHQv2-Cats~\cite{choi2020stargan} contains 5065 cat images of different types.
We randomly select 4k images as the training set and the rest as the test set.
\begin{table*}[t]
{
\caption{\textbf{Quantitative comparison.} 
For AFHQv2-Cats~\cite{choi2020stargan}, since there is non-trivial to estimate depth maps for cat faces, we only evaluate on FID and KID$\times100$.
Note that our results of $512^2$ resolution are synthesized by a super-resolution module and we did not apply the same super-resolution operation to the depth map. Thus, the depth accuracy at this resolution is not available.
The \textbf{bold} highlights the best results among methods requiring single-view images only.
Approaches labeled in \textcolor{gray}{gray} necessitate the availability of multi-view training data.
Legend: * –requires multi-view training data and test time optimization. 
}
    \label{tab: Quantitative comparison}
    \resizebox{0.9\linewidth}{!}{
    \begin{tabular}{l c c c c c c c c c c c c c}
    \toprule
    \multirow{2}{*}{Method} & \multicolumn{4}{c}{FFHQ~\cite{karras2019style}} & & \multicolumn{4}{c}{CelebAMask-HQ~\cite{lee2020maskgan}} &  &\multicolumn{2}{c}{AFHQv2-Cats~\cite{choi2020stargan}}\\
    & FID($\downarrow$) & KID($\downarrow$) & Depth($\downarrow$) & ID($\uparrow$) & & FID($\downarrow$) & KID($\downarrow$) & Depth($\downarrow$) & ID($\uparrow$) & & FID($\downarrow$) & KID($\downarrow$) \\
    \midrule
    Pix2NeRF $64^2$ ~\cite{cai2022pix2nerf}  & 32.44            & 2.37          & 0.40          & 0.25 & & 89.79         & 12.22        & 0.38          & 0.19           & & 25.34           & 1.00\\
    \sexyname $64^2$ (Ours)                  & \textbf{26.04}   & \textbf{2.09} & \textbf{0.35} & \textbf{0.43} & &\textbf{75.76} & 10.48        & \textbf{0.32} & \textbf{0.37}           & & \textbf{18.64}  & \textbf{0.73}\\
    Pix2NeRF $512^2$ ~\cite{cai2022pix2nerf} &  75.04           & 5.97          &  0.41         & 0.20 & & 118.92        & 13.08        & 0.38          & 0.15           & &  50.55          &3.33  \\
    \sexyname $512^2$ (Ours)                 &  40.24           & 2.72          & -             & 0.36 & & 78.38         & \textbf{8.68}& -             & 0.31           & & 21.78           & 1.00\\
    \bottomrule
    \end{tabular}
    }
    \resizebox{0.9\linewidth}{!}{
    \begin{tabular}{l c c c c c c c c c c c }
    \toprule
    \multirow{2}{*}{Method} & \multicolumn{3}{c}{ShapeNet Chairs~\cite{chang2015shapenet}} & & \multicolumn{3}{c}{ShapeNet Cars~\cite{chang2015shapenet}} & & \multicolumn{3}{c}{Average}\\
    & SSIM ($\uparrow$) & PSNR ($\uparrow$) & LPIPS ($\downarrow$)& & SSIM ($\uparrow$) & PSNR ($\uparrow$) & LPIPS ($\downarrow$)&& SSIM ($\uparrow$) & PSNR ($\uparrow$) & LPIPS ($\downarrow$)\\
    \midrule
    \textcolor{gray}{ENR $128^2$~\cite{dupont2020equivariant}}   & \textcolor{gray}{0.91}                & \textcolor{gray}{22.83}             & \textcolor{gray}{0.10}                & &\textcolor{gray}{0.90}          &\textcolor{gray}{22.26}&\textcolor{gray}{0.13}           & &\textcolor{gray}{0.91}             &\textcolor{gray}{22.55}         &\textcolor{gray}{0.12} \\
    \textcolor{gray}{SRN $128^2$~\cite{sitzmann2019scene}}       & \textcolor{gray}{0.89}                & \textcolor{gray}{22.89}             & \textcolor{gray}{0.10}                & &\textcolor{gray}{0.89}          &\textcolor{gray}{22.25}&\textcolor{gray}{0.13}           & &\textcolor{gray}{0.89}             &\textcolor{gray}{22.57}         &\textcolor{gray}{0.12} \\
    \textcolor{gray}{PixelNeRF $128^2$~\cite{cai2022pix2nerf}}   & \textcolor{gray}{0.91}                & \textcolor{gray}{23.72}             & \textcolor{gray}{0.10}                & &\textcolor{gray}{0.90}          &\textcolor{gray}{23.17}&\textcolor{gray}{0.15}           & &\textcolor{gray}{0.91}             &\textcolor{gray}{23.45}         &\textcolor{gray}{0.13}\\
    \textcolor{gray}{CodeNeRF* $128^2$~\cite{jang2021codenerf}}  &\textcolor{gray}{0.90}                 &\textcolor{gray}{23.66}              & \textcolor{gray}{0.11}                & &\textcolor{gray}{0.91}&\textcolor{gray}{23.80}       &\textcolor{gray}{0.12}  & &\textcolor{gray}{0.91}      &\textcolor{gray}{23.73}&\textcolor{gray}{0.12}\\
    \textcolor{gray}{VisionNeRF* $128^2$~\cite{lin2023vision}}   &\textcolor{gray}{0.93}        &\textcolor{gray}{24.48}     & \textcolor{gray}{0.08}       & &\textcolor{gray}{0.91} &\textcolor{gray}{22.28}&\textcolor{gray}{0.08}  & &\textcolor{gray}{0.92}    &\textcolor{gray}{23.37}         &\textcolor{gray}{0.08}\\
    \midrule
    Pix2NeRF $64^2$~\cite{cai2022pix2nerf}      & 0.80                & 18.13             & 0.12                & &0.73             &16.57             &0.19             & &0.77             &17.35&0.16\\
    \sexyname $64^2$ (Ours)                     & \textbf{0.88}    & \textbf{22.31} & \textbf{0.07}    & &\textbf{0.86} &\textbf{21.03} &\textbf{0.10} & &\textbf{0.87} &\textbf{21.67}&\textbf{0.09}\\
    Pix2NeRF $128^2$~\cite{cai2022pix2nerf}     & 0.83                & 17.73             & 0.12                & &0.78             &16.24             &0.20             & &0.81             &16.99&0.16\\
    \sexyname $128^2$ (Ours)                    & \textbf{0.88}    & 20.29             & 0.08                & &\textbf{0.86} &19.44             &0.11             & &\textbf{0.87} &19.87            &0.10\\
    \bottomrule
    \end{tabular}
    }
}
\centering
\end{table*}%

\paragraph{Evaluation metrics.}
\label{sec: Evaluation Metrics}
 Following Pix2NeRF~\cite{cai2022pix2nerf}, we report Frechet-Inception Distance~(\ie~FID)~\cite{heusel2017gans} and Kernel-Inception Distance~(\ie~KID)~\cite{binkowski2018demystifying} for novel view images.
We also use Depth accuracy~(\ie~Depth) to measure the depth quality.
Specifically, we evaluate depth quality by calculating MSE loss against pseudo-ground-truth depth maps estimated from test set images by Deep3DFaceReconstruction~\cite{deng2019accurate, LI2024106275} and our generated depth maps.
We assess multi-view consistency (ID) by calculating the mean Arcface~\cite{deng2019arcface} cosine similarity score between the input images and the corresponding novel views.
In the ablation study, to assess the quality of the generated images, we employ Structural Similarity (SSIM)~\cite{wang2004image}.
\vspace{-10pt}
\paragraph{Implementation details.}
We use two pre-trained models of EG3D~\cite{chan2022efficient} trained on FFHQ~\cite{karras2019style} and AFHQv2-Cats~\cite{choi2020stargan} respectively to synthesize multi-view data.
Specifically, we generate 60k triplets of multi-view data for FFHQ~\cite{karras2019style} and 4k for AFHQv2-Cats~\cite{choi2020stargan}.
We adopt the same pre-processing strategy as~\cite{chan2022efficient}.
All images are aligned and processed into size $512^2$.
Note that we use a super-resolution module to generate a promote a low-resolution image~(\ie~$64^2$) to a high resolution~(\ie~$512^2$).
Please refer to our appendix for more implementation details.

\begin{table}[t]
{
    \caption{\textbf{Comparison of inference cost with PTI~\cite{roich2021pivotal}.}
    We fine-tune an EG3D model to fit a single-view image until it reconstructs the same level of details as ours~(\ie~LPIPS:~0.32). 
    Then, we use this fine-tuned model to synthesize four novel views and compare the time taken at each stage with our model on an RTX A800 GPU.
    }
    \label{tab:  pti}
    \resizebox{1.0\linewidth}{!}{
    \begin{tabular}{l | c | c c c c c }
    \toprule
    Method  & Inference Cost ($\downarrow$)  & Depth ($\downarrow$) & Depth$^\dagger$ ($\downarrow$) & FID ($\downarrow$) & KID ($\downarrow$)& ID ($\uparrow$) \\
    \midrule
    PTI\cite{roich2021pivotal} &76.9s &0.37  &0.55 &35.53  &2.30  &0.36 \\
    Ours &1.1s  &0.35  &0.53 &40.24  &2.72  &0.35 \\
    \bottomrule
    \end{tabular}
    }
}
\centering
\end{table}%


\subsection{Comparison with State-of-the-art Methods}

\begin{figure}[t]
    \centering
    \vspace{-10pt}
    \includegraphics[width=1.0\linewidth]{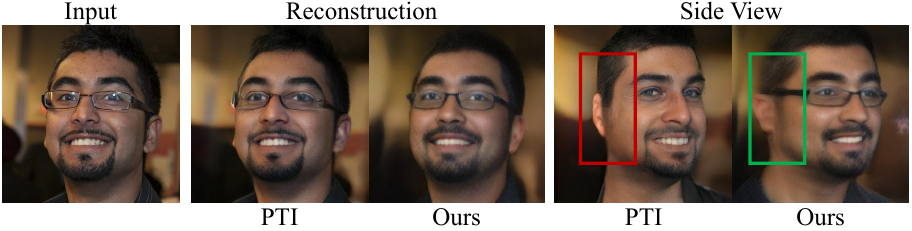}
    \vspace{-20pt}
    \caption{\hzx{\textbf{Qualitative comparisons with PTI~\cite{roich2021pivotal}}.}}
    \vspace{-10pt}
    \label{fig: qualitative comparisons with pti}
\end{figure}

\paragraph{Quantitative comparison.} \label{Quantitative Comparison.}
We compare our method against the state-of-the-art method Pix2NeRF~\cite{cai2022pix2nerf} for novel-view synthesis from a single image with real-world single-view datasets.
Note that we use scripts provided by the authors of Pix2NeRF~\cite{cai2022pix2nerf} to train with FFHQ~\cite{karras2019style} dataset and AFHQv2-Cats~\cite{choi2020stargan} dataset.
Since AFHQv2-Cats contains a relatively small amount of data, we train for 150k iterations with a batch size of 48 for Pix2NeRF~\cite{cai2022pix2nerf} until convergence.
Tab.~\ref{tab: Quantitative comparison} provides quantitative metrics comparing the proposed approach against Pix2NeRF~\cite{cai2022pix2nerf}.
Our model demonstrates better results in terms of all metrics across all datasets.
Moreover, our model is capable of generating higher-resolution images without compromising on efficiency, whereas Pix2NeRF requires significantly more time to achieve comparable results~\cite{chan2022efficient}.
In other words, our model can generate novel views with more realistic appearances and precise shapes from a single-view image.
Notably, Pix2NeRF fails to synthesize novel views on AFHQv2-Cats~\cite{choi2020stargan}, because most of the cat faces in this dataset are facing the camera, resulting in limited geometry information.
In contrast, our model derives advantages from training with a collection of synthetic multi-view images and a depth-aware discriminator, which facilitates explicit 3D supervision. 
Consequently, our model effectively learns a robust geometry prior, even in such challenging scenarios.
\vspace{-10pt}

\begin{figure}[t]
    \centering
    \includegraphics[width=0.93\linewidth]{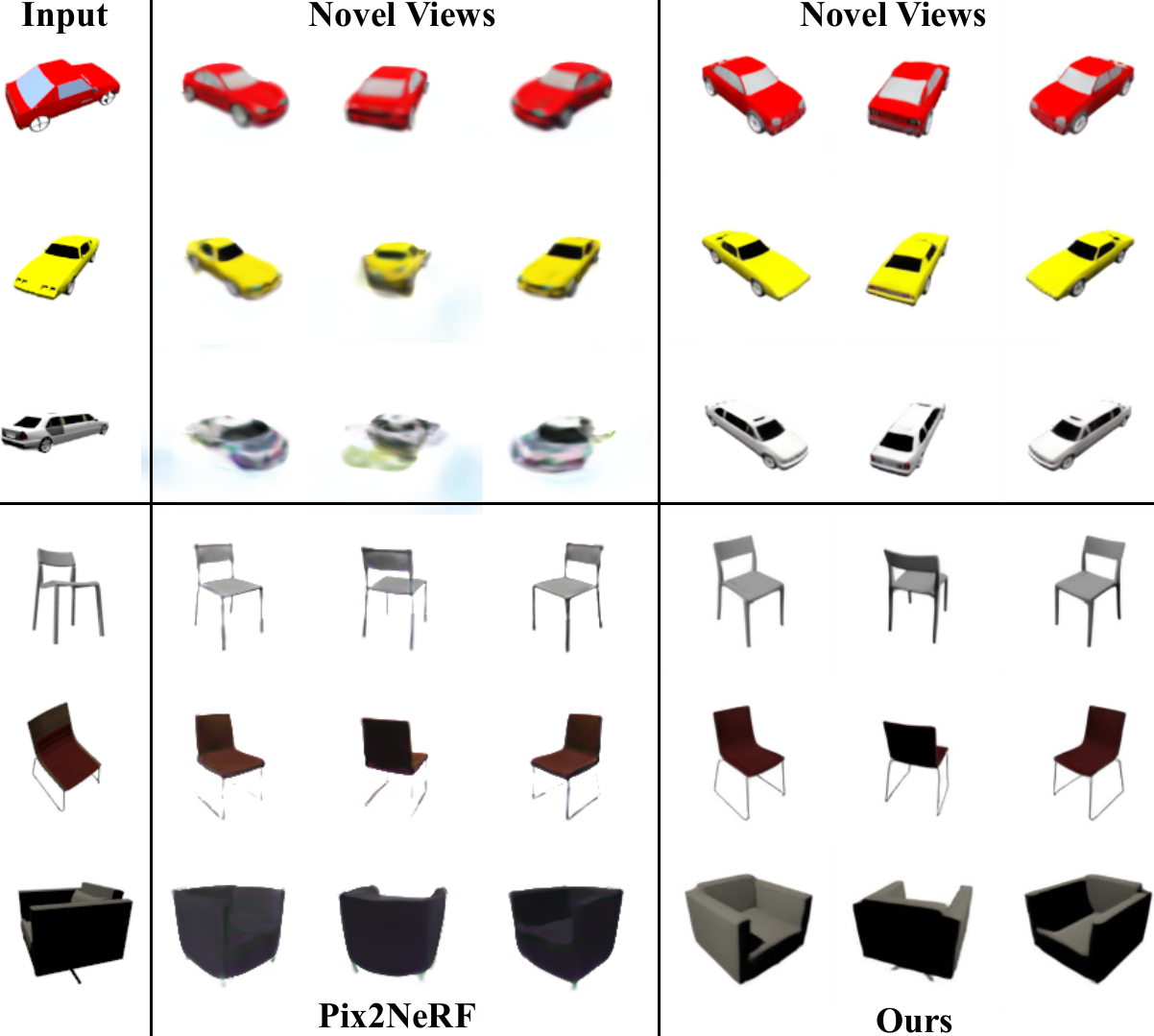}
    \caption{\textbf{Qualitative comparisons with Pix2NeRF~\cite{cai2022pix2nerf} on  ShapeNet Cars \& Chairs~\cite{chang2015shapenet, sitzmann2019scene}.}}
    \label{fig: qualitative comparisons on multi-view dataset}
    \vspace{-15pt}
\end{figure}

\paragraph{Qualitative comparison.}
Fig.~\ref{fig: Qualitative Comparison} presents results generated by our method and Pix2NeRF~\cite{cai2022pix2nerf} on FFHQ~\cite{karras2019style}, CelebAMask-HQ~\cite{lee2020maskgan}, and AFHQv2-Cats~\cite{choi2020stargan}.
Our method can synthesize high-quality novel views even with a single image as reference, yet existing few-shot NeRF methods can not train on these single-view datasets without multi-view image pairs.
Compared to Pix2NeRF~\cite{cai2022pix2nerf}, our methods can generate more realistic results while preserving a more similar identity to the reference images.
Thanks to the depth-aware discriminator, our method excels in producing high-quality results even under extreme camera poses~(see the bottom row of the leftmost column in Fig.~\ref{fig: Qualitative Comparison}).
Meanwhile, our method can also learn geometry priors from AFHQv2-Cats~\cite{choi2020stargan} which contains a limited range of poses while Pix2NeRF~\cite{cai2022pix2nerf} fails~(see the rightmost column in Fig.~\ref{fig: Qualitative Comparison}).
We also provide more geometry visualization results, please refer to our appendix for more details. 
\vspace{-10pt}

\begin{figure}[t]
    \centering
    \includegraphics[width=0.9\columnwidth]{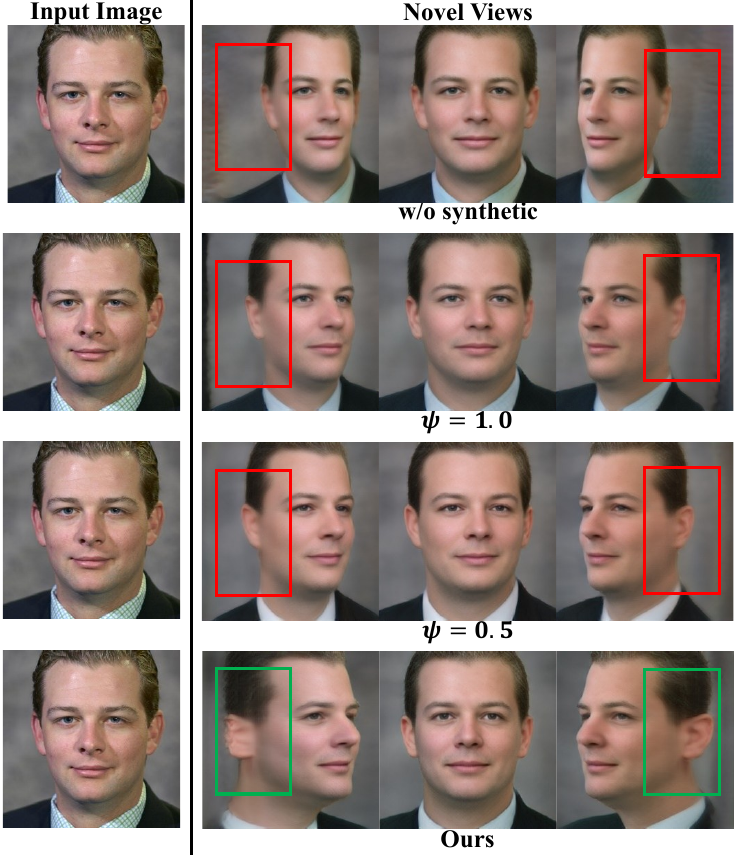}\label{fig: w/o mixed}
    \vspace{-10pt}
    \caption{\textbf{Ablation study.}
    Without incorporating truncation method ~(~\ie~$\psi=1.0$) and depth-aware training, our model fails to generate results with realistic geometry~(see the \textcolor{red}{red} boxes in the figure).
    }
    \label{fig: Ablation Studies of GMVS}
    \vspace{-15pt}
\end{figure}




\paragraph{Comparison on multi-view datasets.}
Additionally, we performed experiments on the training set of ShapeNet Cars \& Chairs~\cite{chang2015shapenet, sitzmann2019scene}, which includes uniformly distributed camera poses around a sphere.
Following the experimental setting of Pix2NeRF~\cite{cai2022pix2nerf}, we filtered the training set for both datasets to only include the upper hemisphere and evaluate the test split.
Since EG3D~\cite{chan2022efficient} does not include an evaluation on ShapeNet Chairs, we first train an EG3D model using the same settings as those used for training ShapeNet Cars.
We evaluate the performance using well-established image quality metrics commonly employed in novel view synthesis tasks, including pixel-level measures such as SSIM and PSNR, as well as a feature-level metric called LPIPS.

As depicted in Tab.~\ref{tab: Quantitative comparison}, our method consistently outperforms Pix2NeRF~\cite{cai2022pix2nerf} in terms of all metrics.
Simultaneously, our method remains competitive with other approaches that rely on multi-view training data, while our method does not employ multi-view supervision on the ShapeNet datasets~\cite{chang2015shapenet}.
Fig.~\ref{fig: qualitative comparisons on multi-view dataset} demonstrates our method's superior accuracy in shape and texture predictions when compared to Pix2NeRF~\cite{cai2022pix2nerf}.
\vspace{-10pt}

\paragraph{Comparison with GAN inversion method.}
We employ Pivotal Tuning Inversion~(PTI)~\cite{roich2021pivotal} to fine-tune an EG3D model, enabling its adaptation to a single-view image.
The fine-tuning process continues until the LPIPS~\cite{zhang2018unreasonable} loss matches that of our method. 
Subsequently, we compare the time taken by our method with that of the fine-tuning-based approach.
As shown in Tab.~\ref{tab: pti}, our method demonstrates significantly faster than the GAN inversion method.
\hzx{
Moreover, our method outperforms PTI in terms of geometry quality (Depth and Depth$^\dagger$) and inference cost while being competitive in image quality (FID, KID, and ID).
Visual results also show the quality of another view produced by PTI exhibits instability~(see Fig.~\ref{fig: qualitative comparisons with pti}).
}

\subsection{Ablation Study}
\begin{table}[t]
{
    \caption{\textbf{Quantitative results of ablation studies.}
    \label{tab: Ablation Studies}
    \hzx{
    The row without truncation ratio means training without synthetic data.} 
    The \textbf{bold} numbers highlight the best results.
    $^\dagger$ Evaluated on side faces.} 
    \resizebox{1.0\linewidth}{!}{
    \begin{tabular}{c c c | c c c c c c}
    \toprule
    
    Trunc. Ratio     &Real Img. & $\mathcal{D}_{g}$  & FID ($\downarrow$) & KID ($\downarrow$) & Depth ($\downarrow$) & Depth$^\dagger$ ($\downarrow$) & ID ($\uparrow$) & SSIM ($\uparrow$)\\
    \midrule
      \xmark   & \cmark   &\xmark   &\textbf{33.13} &\textbf{2.22}  &0.42            &0.83             &\textbf{0.43} &\textbf{0.66}   \\
           1.0 & \cmark   &\xmark   &34.72         &2.42            &0.37            &0.64             &0.38          &0.65            \\
           0.5 & \cmark   &\xmark   &40.13         &2.64            &0.35            &0.59             &0.35          &0.64            \\
           0.5~(Ours) & \cmark   &\cmark   &40.24         &2.72            &\textbf{0.35}   &\textbf{0.53}    &0.35          &0.63            \\
    \bottomrule
    \end{tabular}
    }
}
\centering
\vspace{-5pt}
\end{table}%
In this part, we provide more experiments and analysis of our proposed modules to verify their effectiveness.
For simplicity, we only conduct experiments on FFHQ~\cite{karras2019style}, and all the depth accuracy is evaluated at a resolution of $64^2$ and the others are evaluated at $512^2$.
\hzx{The data presented in Tab.~\ref{tab: Ablation Studies} correspond to the cases in Fig.~\ref{fig: Ablation Studies of GMVS}, which visualizes the results of different training settings.}

\vspace{-10pt}
\paragraph{Impact of geometry-guided multi-view synthesis scheme.}
 We train our model using synthetic data generated with varying truncation ratios to assess the effectiveness of our Geometry-guided Multi-View Synthesis scheme.
 \hzx{As shown in Tab.~\ref{tab: Ablation Studies}~(the first two rows), our model with synthetic data obtains better depths and competitive identities (FID, KID, ID, and SSIM) compared to the model without synthetic data.
 When decreasing the truncation ratio (1.0 $\rightarrow$ 0.5), the depth quality can be further improved.
 }
Particularly, we can observe a significant drop in the depth accuracy when applying a truncation ratio of 1.0, which severely harms the realism of the generated scenes~(see the bottom row in Fig.~\ref{fig: Ablation Studies of GMVS}).
In essence, if the truncation trick is not taken, it is challenging for our model to generate realistic results.


\vspace{-12pt}
\paragraph{Impact of depth-aware discriminator.}
To verify the effectiveness of our depth-aware discriminator, we train a model without $\mathcal{D}_{g}$.
We can see from Fig.~\ref{fig: Ablation Studies of GMVS} that the model learns a degenerate solution where the human head appears flattened and sunk into the background.
This phenomenon typically occurs when one of the ears is not visible in the input image.
As shown in Tab.~\ref{tab: Ablation Studies}, the removal of $\mathcal{D}_{g}$ yields to a slight improvement in image quality metric~(\ie~SSIM:~$0.63\rightarrow0.64$). 
This is primarily because the addition of a discriminator naturally brings some disturbance to our training process.

As $\mathcal{D}_{g}$ is directly applied to depth maps, depth accuracy is the most important metric for verifying its effectiveness.
From the fifth column in Tab.~\ref{tab: Ablation Studies}, we can observe that the depth accuracy remains almost unchanged without $\mathcal{D}_{g}$~(\ie~$0.35\rightarrow0.35$).
However, the sixth column in Tab.~\ref{tab: Ablation Studies} reveals a significant drop in depth accuracy~(\ie~$0.53\rightarrow0.59$) when faces are turned to the side.
In other words, $\mathcal{D}_g$ plays a crucial role in distinguishing unnatural depth maps and contributes to achieving a more realistic geometry from various viewpoints.

\begin{figure}[t]
    \centering
    \includegraphics[width=0.80\linewidth]{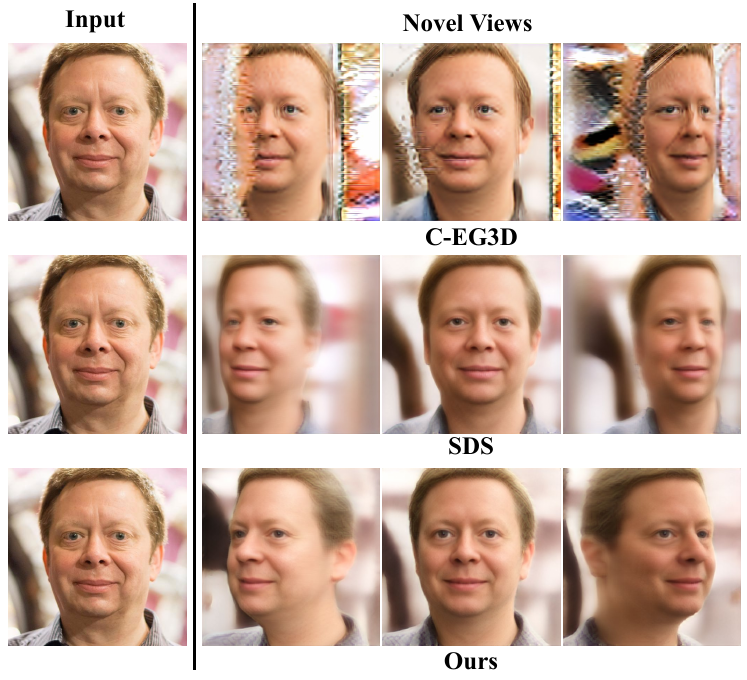}
    \vspace{-10pt}
    \caption{\textbf{Further comparisons with two intuitive methods.} We compare our \sexyname with two intuitive methods to further verify the effectiveness of our method.}
    \label{fig: intuitive methods}
    \vspace{-15pt}
\end{figure}

\subsection{Further Discussion}
We conduct two intuitive comparisons to further validate the effectiveness of our method: 1) Conditional EG3D~(denoted as C-EG3D). 
We condition the EG3D~\cite{chan2022efficient} model on an input image and performed novel view synthesis using only a single-view reconstruction loss and the GAN loss as used in the original EG3D paper~\cite{chan2022efficient}. 
However, as shown in Fig.~\ref{fig: intuitive methods}, this approach fails to capture 3D information from the input image. 
It also exhibits artifacts when presented with novel viewpoints. 
We recognize that this is a challenging one-to-many problem, as it is non-trivial for models to learn geometry without explicit multi-view supervision or accurate depth information.
2) Simple depth supervision~(denoted as SDS). 
We use a state-of-the-art depth estimation model, such as MiDaS~\cite{Ranftl2022}, to generate depth maps for images in the FFHQ dataset~\cite{karras2019style}. 
After scaling and adjusting these depth maps, we employ them for basic depth supervision along with a single-view reconstruction loss, without incorporating our proposed geometry-guided multi-view synthesis scheme. 
In Fig.~\ref{fig: intuitive methods}, this approach fails to produce accurate geometry, resulting in flat, plane-like facial reconstructions. 
The limitations stem from the challenge of MiDaS~\cite{Ranftl2022} in estimating fine-grained facial details, such as the nose, mouth, and eyes.

\section{Conclusion}
In this work, we propose the ~\sexyname, a single-shot novel view synthesis method designed for high-fidelity novel view synthesis using only real-world single-view images. 
\sexyname seeks to enhance geometry priors into a NeRF model through two stages: Geometry-guided Multi-View Synthesis (GMVS) and Depth-aware Training (DaT). GMVS leverages an off-the-shelf 3D GAN model 
to synthesize multi-view data, enhanced with a truncation method for improved geometry quality. DaT further refines the NeRF model by incorporating a depth-aware discriminator, guiding the learning process through depth maps. Our proposed method is evaluated extensively on multiple real-world datasets and the experimental results demonstrate its effectiveness.

\section{Acknowledgement}
 This work was partially supported by National Natural Science Foundation of China (NSFC) 62072190, Program for Guangdong Introducing Innovative and Entrepreneurial Teams 2017ZT07X183, and TCL Science and Technology Innovation Fund.

%% file: sec/X_suppl.tex
\setcounter{section}{0}
\setcounter{equation}{0}
\setcounter{table}{0}
\setcounter{figure}{0}
\renewcommand\thesection{\Alph{section}}
\renewcommand{\thetable}{\Alph{table}}
\renewcommand{\thefigure}{\Alph{figure}}

    

In the appendix, we provide more details and more experimental results of the proposed 
We organize the appendix into the following sections. 

\begin{itemize}

    \item In section~\ref{sec: Preliminary of Neural Radiance Fields}, we depict the preliminary of NeRF.
    
    \item In section~\ref{sec: implementation details}, we provide more implementation details.
    
    \item In section~\ref{sec: additional results}, we show more qualitative and quantitative results.
    
    \item In section~\ref{More discussions}, we provide more discussions about the potential limitations of our method and the difference with concurrent works. 
\end{itemize}

\section{Preliminary of Neural Radiance Fields}
\label{sec: Preliminary of Neural Radiance Fields}
NeRF~\cite{mildenhall2020nerf} aims to synthesize novel views of complex scenes from sparse input views. By querying 5D coordinates along a camera ray and leveraging volume rendering technique, NeRF generates the color of an image pixel that intersects with the ray.  Specifically, for a given pixel coordinate $\mathbf{x} \in \mathbb{R}^2$ and camera parameters $\mathbf{P}$ of an image, we acquire a camera ray $\mathbf{r}(t)=\mathbf{o}+t\mathbf{d} $ where $\mathbf{o}$ is camera center, $\mathbf{d} \in \mathbb{S}^2$ denotes view direction calculated with $\mathbf{o}$, $\mathbf{x}$ and $\mathbf{P}$.
Here, the procedure of acquiring rendered image color can be defined as the following equations with near and far bounds $t_n$ and $t_f$ regarding $\mathbf{r}$:
\begin{equation}
    \begin{aligned}
        \mathbf{\hat{C}}(\mathbf{r})=\int_{t_{n}}^{t_{f}}T(t)\sigma(\mathbf{r}(t))\mathbf{c}(\mathbf{r}(t), \mathbf{d})dt,\\
        T(t)={\rm exp}(-\int_{t_{n}}^t\sigma(\mathbf{r}(s))ds), 
     \label{continous volume rendering}
     \end{aligned}
\end{equation}
where the density $\sigma(x)$ is the probability that the ray terminates at a particle. $T(t)$ denotes the probability the ray $\mathbf{r}$ travels from $t_n$ to $t$ without hitting any particle.
To numerically estimate the continuous integral (Eq.~\ref{continous volume rendering}), NeRF samples particles along the continuous camera ray $\mathbf{r}(t)$ with a stratified sampling strategy in which $\mathbf{r}(t)$ is evenly partitioned into $n$ bins. By querying the position and direction of each particle with a multi-layer perception (MLP), we obtain the color and density of each particle. Using Eq.~\ref{eqn:nerf_color_depth}, we accumulate the color and density of particles along a ray for the pixel color of the rendered image:
\begin{equation}
    \begin{split}
    \mathbf{\hat{C}(\mathbf{r})}=\sum_{i=1}^{N_s}\tau_i\alpha_i\mathbf{c}(\mathbf{r}(t_i)),~
    ~\mathbf{\hat{D}(\mathbf{r})}=\sum_{i=1}^{N_s}\tau_i\alpha_{i}{z}_{i},\\
    {\rm where}~\tau_i=\prod_{j=1}^{i-1}(1-\alpha_j),~~\alpha_i=1-e^{-\sigma(\mathbf{r}(t_i))\delta_i},
    \end{split}
    \label{eqn:nerf_color_depth}
\end{equation}
where $\tau_i$ denotes the accumulated transmittance along the ray from
the $t_n$ to $t_f$ and $\delta_i= t_{i+1}-t_i $ is the distance of two adjacent particles.
$~\mathbf{\hat{D}(\mathbf{r})}$ is depth value of the rendered image and ${z}_{i}$ denotes the depth of the $i_{th}$ particle in the stratified $r(t)$. Since estimating $\mathbf{\hat{C}(\mathbf{r})}$ from  $\mathbf{c}(\mathbf{r})$ and $\sigma(\mathbf{r}(t_i))$ is differentiable, NeRF is optimized via minimizing the MSE loss between $\hat{\mathbf{C}}(\mathbf{r})$ and the ground truth color $\mathbf{C}(\mathbf{r})$ via equation $\mathcal{L}_{nerf}=\vert \vert \hat{\mathbf{C}}(\mathbf{r}) - \mathbf{C}(\mathbf{r}) \vert \vert _2 ^2 $. 
\vspace{-5pt}
\begin{figure}[t]
    \centering
    \includegraphics[width=0.95\linewidth]{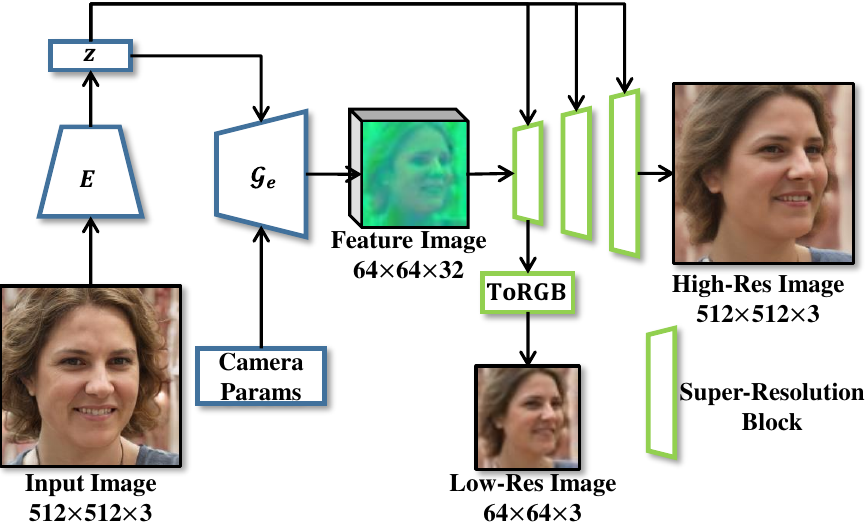}
    \caption{\textbf{More details of our network architecture.}
    Our \sexyname consists of a scene encoder ${E}$, an EG3D backbone $\mathcal{G}_{e}$ and three super-resolution blocks.}
    \vspace{-10pt}
    \label{fig: network architeture}
\end{figure}

\section{More implementation Details} 
\label{sec: implementation details}

\subsection{More details of our network architecture}
The network architecture of \sexyname is depicted in Fig.~\ref{fig: network architeture}. The structure of scene encoder ${E}$ is ResNeXt~\cite{xie2017aggregated} and borrow from \cite{zhou2021pose}.
The NeRF-based generator $\mathcal{G}_n$ consists of an EG3D backbone $\mathcal{G}_e$ and a super-resolution module.
The structure of the EG3D-backbone $\mathcal{G}_e$ and depth-aware discriminator $\mathcal{D}_{g}$ are borrowed from~\cite{chan2022efficient}.
Differently, to capture more information, we increase the latent code dimension of $\mathcal{G}_e$ to 5120.
The super-resolution module includes three super-resolution blocks, which are the same blocks used in $\mathcal{G}_e$.
All these modules are trained from scratch together.

\subsection{More experimental details}
\hzx{All experiments are conducted on PyTorch~\cite{paszke2019pytorch} with 2 80GB RTX A800 GPUs.
We use Adam~\cite{kingma2014adam} with $\beta_{1}=0.9$, $\beta_{2}=0.999$ for $\mathcal{E}$ and $\mathcal{G}_{n}$, and $\beta_{1}=0$, $\beta_{2}=0.99$ for $\mathcal{D}_{g}$.
We set the learning rate as 1e-03 for the generator and 8e-06 for the discriminator.
For the hyperparameter $\lambda_{g}$, we empirically set it to 1.2.
We train our model with FFHQ~\cite{karras2019style} for 4000k images with batch size 24 and for 2000k images with AFHQv2-Cats~\cite{choi2020stargan}.}

In comparisons on ShapeNet datasets, since Pix2NeRF~\cite{cai2022pix2nerf} does not include an evaluation on the ShapeNet Cars~\cite{chang2015shapenet,sitzmann2019scene}, we first train a Pix2NeRF model using the official code.
We generate 62,000 synthetic images for ShapeNet Cars and 140,000 for ShapeNet Chairs.
Subsequently, we incorporate the synthetic images with the ShapeNet datasets to train our model.
During the evaluation phase, for each category, we randomly select one image as the input and ten images as ground truths.

\section{Additional Results}
\label{sec: additional results}

\subsection{Training without real-world images}
\begin{figure}[t]
    \centering
    \includegraphics[width=0.85\linewidth]{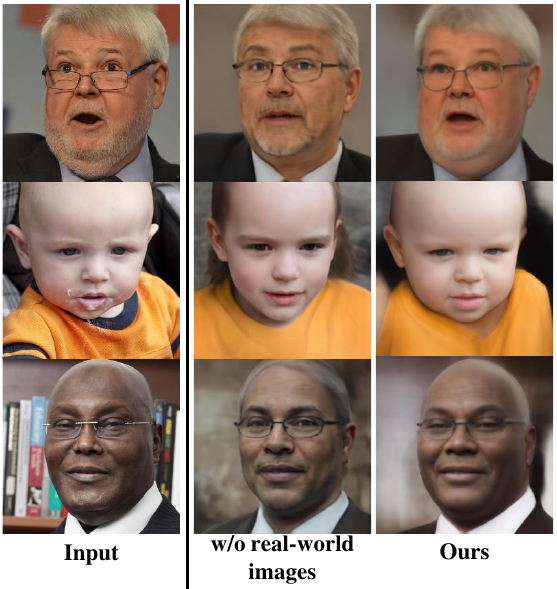}
    \caption{\textbf{Ablation study of incorporating real-world images.}
    Without incorporating real-world images, our model generates results with poor similarity to the input images.}
    \label{fig: without real images}
    \centering
\end{figure}

Since we trade off the diversity and geometry quality in synthetic multi-view data through a truncation ratio of 0.5, it is important to incorporate real-world images to provide diverse appearance priors.
We train our model without incorporating real-world images to verify its effectiveness.
As seen from Fig.~\ref{fig: without real images} and Tab.~\ref{tab: Quantitative results of training without real-world images}, our model produces results with poor similarity to the reference images and all the evaluation metrics decrease compared to our full model.

\subsection{More comparison with  HeadNeRF}
\begin{figure}[t]
    \centering
    \includegraphics[width=0.9\linewidth]{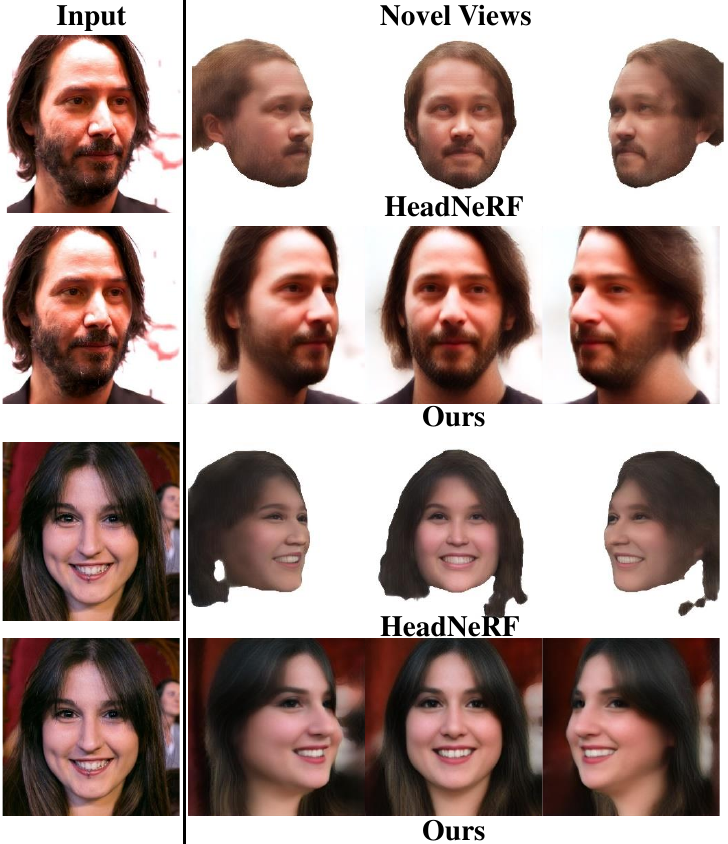}
    \vspace{-10pt}
    \caption{\textbf{Qualitative comparison with HeadNeRF~\cite{hong2021headnerf}.}
    Compared to HeadNeRF~\cite{hong2021headnerf}, our method synthesizes more ID-preserving and realistic novel views.}
    \label{fig: Qualitative comparison with HeadNeRF}
\end{figure}

We compare our method with another single-shot novel view synthesis method named HeadNeRF~\cite{hong2021headnerf}, which introduces a NeRF-based parametric head model to synthesize controllable 3D faces from single-view images.
As shown in Fig.~\ref{fig: Qualitative comparison with HeadNeRF}, HeadNeRF~\cite{hong2021headnerf} is incapable of generating ID-preserving results but inherits some artifacts due to the limitation of the  3D Morphable Models (3DMMs).
Moreover, HeadNeRF~\cite{hong2021headnerf} also requires additional training processes to fit a single image.
 In contrast, our G-NeRF achieves high-fidelity novel view synthesis without any test-time optimization.

\subsection{More qualitative results}
\begin{figure*}
    \centering
    \includegraphics[width=2\columnwidth]{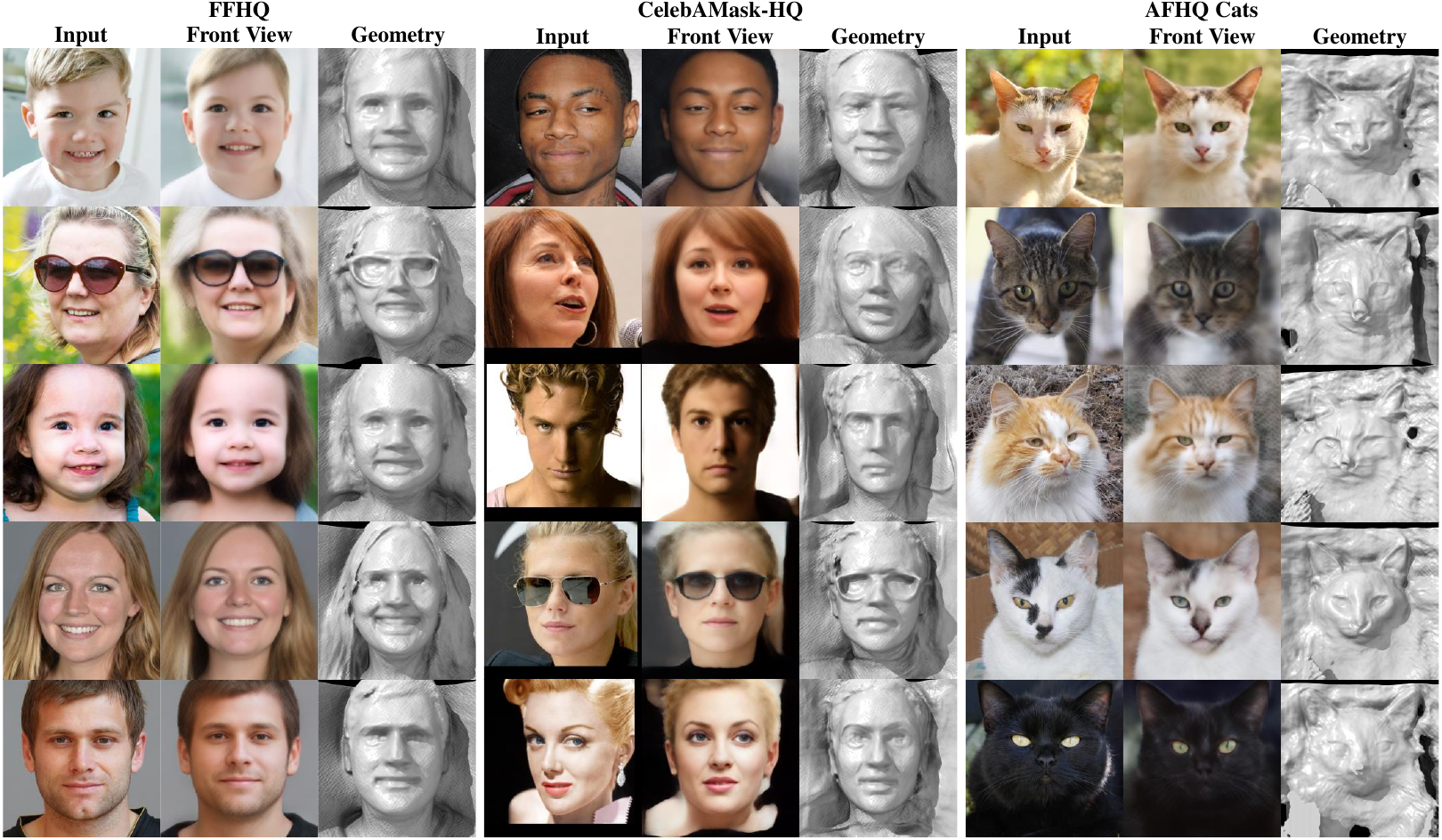}
    \vspace{-10pt}
    \caption{\textbf{More qualitative results.}
        We provide front views and geometry generated with various reference images.
        Our method is capable of synthesizing novel views for diverse input on FFHQ~\cite{karras2019style}, CelebAMask-HQ~\cite{lee2020maskgan}, and AFHQv2-Cats~\cite{choi2020stargan}.
        }
    \label{fig: more results}
    \vspace{-10pt}
\end{figure*}

We present additional qualitative results using various reference images from three datasets, including FFHQ~\cite{karras2019style}, CelebAMask-HQ~\cite{lee2020maskgan} and AFHQv2-Cats~\cite{choi2020stargan}.
Specifically, for each reference image, we generate a front view and estimate its geometry following the approach described in~\cite{chan2022efficient}.
To visualize the geometry results, we use ChimeraX~\cite{goddard2018ucsf}.
As shown in Fig.~\ref{fig: more results}, \sexyname successfully synthesizes novel views with accurate geometry for a wide range of input images.
Notably, our approach can handle inputs with glasses, complex lighting environments, different ages, and varying viewpoints.
Although the geometry of AFHQv2-Cats~\cite{choi2020stargan} exhibits some hole artifacts due to the limited poses available, we still can generate novel views for these cat faces.

\begin{table}[t]
{
    \caption{\textbf{Impact of incorporating real-world images.}
    \label{tab: Quantitative results of training without real-world images}
    The \textbf{bold} numbers highlight the best results.}
    \vspace{-5pt}
    \resizebox{1\linewidth}{!}{
    \begin{tabular}{c c c | c c c }
    \toprule
    Trunc. Ratio     & real-world images & $\mathcal{D}_{g}$  & SSIM ($\uparrow$) & Depth ($\downarrow$) & ID ($\uparrow$) \\
    \midrule
            0.5 & \xmark & \cmark   & 0.55              & 0.37                  & 0.16             \\
            0.5 & \cmark & \cmark   & 0.63     & \textbf{0.35}         & 0.35 \\
    \bottomrule
    \vspace{-15pt}
    \end{tabular}
    }
}
\centering
\end{table}%

\begin{table}[t]
{
    \caption{Quantitative results of different latent code dimensions.
    \label{tab: Ablation Studies of latent code dimensions}
    A larger latent dimension can offer increased capacity for models to capture more information, thereby leading to better performance.}
    \resizebox{1.0\linewidth}{!}{
    \begin{tabular}{l c c c c} 
    \toprule
    Latent code dim.       & LPIPS ($\downarrow$) & Depth ($\downarrow$) & FID ($\downarrow$) & KID ($\downarrow$)\\
    \midrule
    512        & 0.36                 & 0.36                 & 51.68              & 3.68\\
    1024       & 0.35                 & 0.36                 & 49.20              & 3.66\\
    3072       & 0.34                 & 0.37                 & 43.10              & 3.07\\
    5120~(ours)& \textbf{0.33}        & \textbf{0.35}        & 40.24              & 2.72\\
    7168       & \textbf{0.33}        & 0.36                 & \textbf{39.27}     & \textbf{2.63}\\
    \bottomrule
    \end{tabular}
    }
}
\centering
\vspace{-13pt}
\end{table}%


\subsection{Impact of latent code dimension}
The dimension of the latent code has a substantial impact on the overall performance. 
To evaluate the effectiveness of a larger latent code dimension, we conducted experiments using different dimension sizes. 
As illustrated in Tab.~\ref{tab: Ablation Studies of latent code dimensions}, increasing the latent code dimension enhances the model's capability to capture finer details. 
Nevertheless, once the dimension reaches 5120, the incremental benefits of further expansion become negligible. Taking into account GPU memory consumption, we ultimately decided to set the latent code dimension to 5120.

\section{More discussions}
\label{More discussions}

\subsection{More discussion with concurrent work}
In a concurrent study~\cite{trevithick2023}, a pre-trained EG3D model is adopted to synthesize a collection of synthetic images for single-shot novel view synthesis of human faces and cat faces. Although this work demonstrates the effectiveness of synthetic face images, it overlooks the fact that models trained solely on synthetic data are susceptible to a gradual decline in either quality (precision) or diversity (recall)~\cite{alemohammad2023self}.
In contrast, our method tackles this issue by incorporating real-world images into training our model.
Meanwhile, no experimental results were provided on $360^\circ$ datasets such as ShapeNet Cars~\cite{chang2015shapenet, sitzmann2019scene}. Therefore, the performance of the proposed method, when trained on this particular dataset, remains uncertain.

\subsection{Potential limitations}
\begin{figure}[t]
    \centering
    \includegraphics[width=0.95\linewidth]{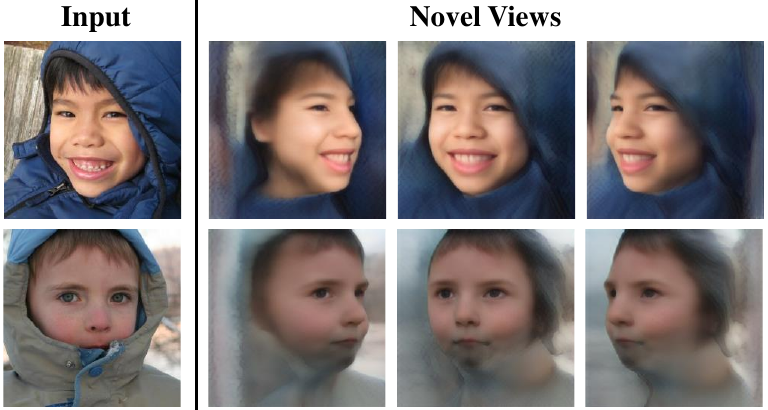 }
    \vspace{-5pt}
    \caption{\textbf{Failure Cases.}
    Our method may encounter failures when faces are occluded by items such as clothing.}
    \vspace{-15pt}
    \label{fig: Failure Cases}
\end{figure}

Our method may encounter failures when faces are occluded by items such as clothing, as illustrated in the first two rows of Fig.~\ref{fig: Failure Cases}.
Unlike faces which often have a similar shape, these irregular occlusions present a challenge for our model to capture a common geometry prior.
As a result, this occlusion leads to a blurred area around the faces.
We will address the above limitation and extend our method to more complex cases like 3D clothed human reconstruction~\cite{yang2024hilo}.
\clearpage

\begin{figure*}
    \centering
    \includegraphics[width=1.0\linewidth]{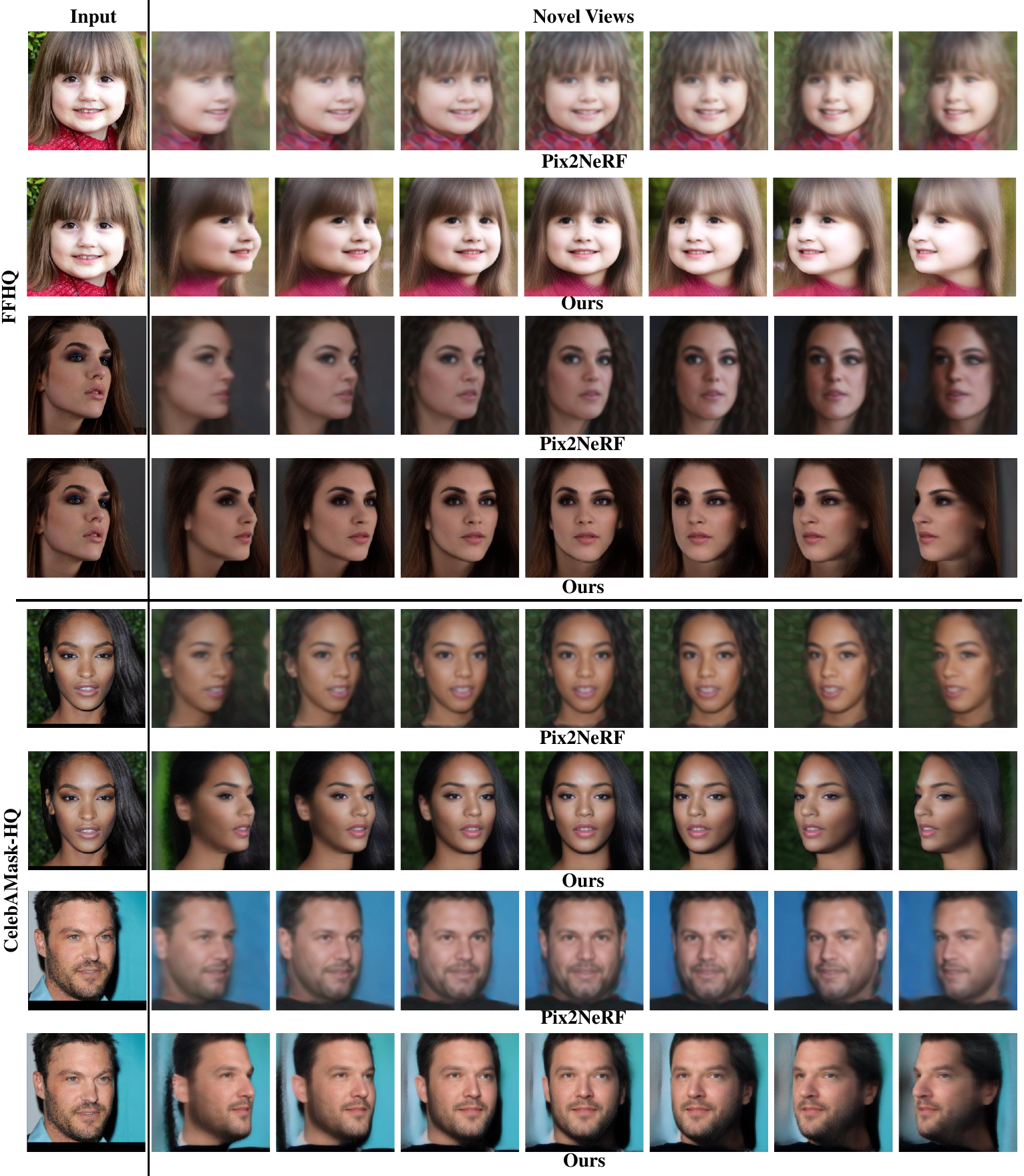  }
    \caption{More qualitative comparisons with Pix2NeRF~\cite{cai2022pix2nerf} on FFHQ~\cite{karras2019style} and CelebAMask-HQ~\cite{lee2020maskgan}.}
    \label{fig: More Comparison swith Pix2NeRF}
\end{figure*}
\clearpage

\begin{figure*}
    \centering
    \includegraphics[width=1.0\linewidth]{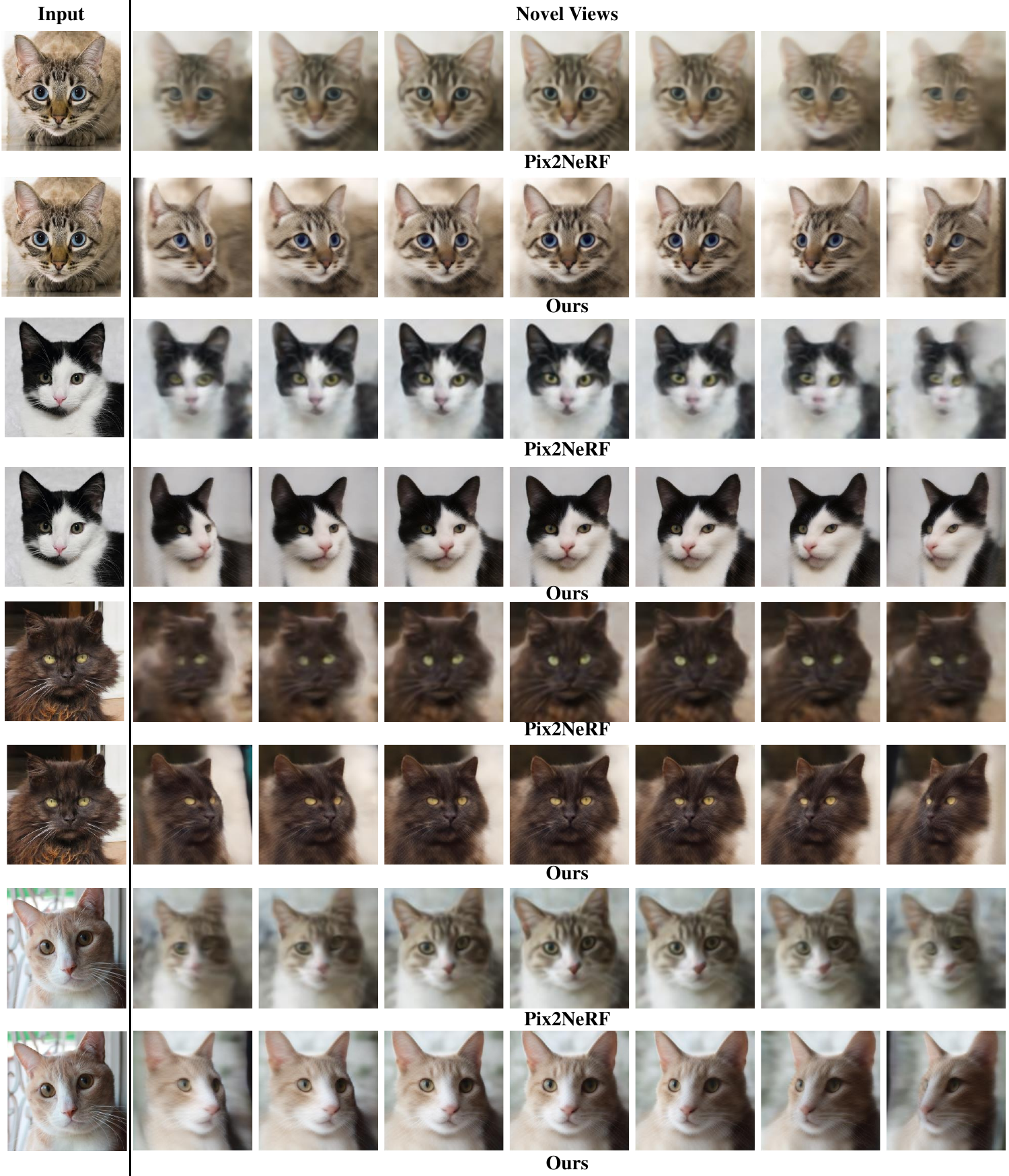  }
    \caption{More qualitative comparisons with Pix2NeRF~\cite{cai2022pix2nerf} on AFHQv2-Cats~\cite{choi2020stargan}.}
    \label{fig: More Comparisons with Pix2NeRF on AFHQV2}
\end{figure*}
\clearpage

\begin{figure*}
    \centering
    \includegraphics[width=1.0\linewidth]{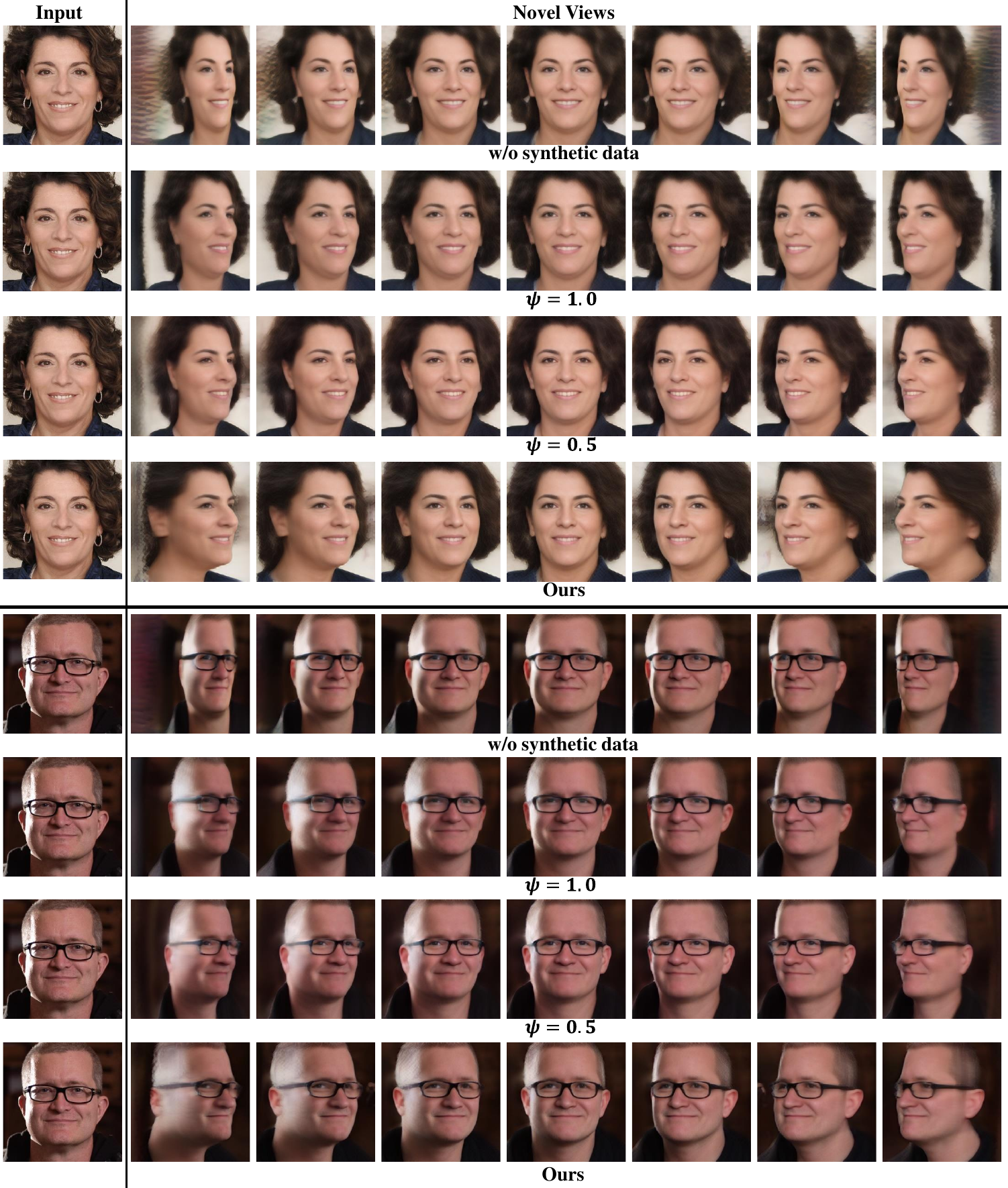  }
    \caption{More qualitative results of ablation studies.}
    \label{fig: More qualitative results of ablation studies}
\end{figure*}

\clearpage